\pdfoutput=1

\documentclass[11pt]{article}

\usepackage[preprint]{acl}

\usepackage{times}
\usepackage{latexsym}

\usepackage[T1]{fontenc}

\usepackage[utf8]{inputenc}

\usepackage{microtype}

\usepackage{inconsolata}

\usepackage{graphicx}

\usepackage{xcolor}

\usepackage{amsmath}

\usepackage{amssymb}
\usepackage{enumitem}

\usepackage{placeins}

\usepackage{booktabs}
\usepackage{multirow}
\usepackage{multicol}
\usepackage{colortbl}
\usepackage{xcolor}
\definecolor{customgray}{gray}{0.9}  %
\definecolor{customblue}{rgb}{0.85,0.9,1}  %
\usepackage{caption}

\title{Improving Preference Extraction In LLMs By Identifying Latent Knowledge Through Classifying Probes}

\author{
 \textbf{Sharan Maiya\textsuperscript{1,2}},
 \textbf{Yinhong Liu\textsuperscript{1}},
 \textbf{Ramit Debnath\textsuperscript{2}},
 \textbf{Anna Korhonen\textsuperscript{1}},
\\
 \textsuperscript{1}Language Technology Lab, University of Cambridge, \\
 \textsuperscript{2}Cambridge Collective Intelligence \& Design Group, University of Cambridge,
\\
 \small{
   \textbf{Correspondence:} \href{mailto:sm2783@cam.ac.uk}{sm2783@cam.ac.uk}
 }
}

\begin{document}
\maketitle
\begin{abstract}
Large Language Models (LLMs) are often used as automated judges to evaluate text, but their effectiveness can be hindered by various unintentional biases. We propose using linear classifying probes, trained by leveraging differences between contrasting pairs of prompts, to directly access LLMs' latent knowledge and extract more accurate preferences. Through extensive experiments using models of varying size from four different families and six diverse datasets assessing text quality evaluation and common sense reasoning, we demonstrate that both supervised and unsupervised probing approaches consistently outperform traditional generation-based judgement while maintaining similar computational costs. These probes generalise under domain shifts and can even outperform finetuned evaluators with the same training data size. Our results suggest linear probing offers an accurate, robust and computationally efficient approach for LLM-as-judge tasks while providing interpretable insights into how models encode judgement-relevant knowledge. Our data and code will be openly released in the future.
\end{abstract}

\section{Introduction}

\begin{figure*}[!ht]
    \centering
    \includegraphics[width=0.48\linewidth]{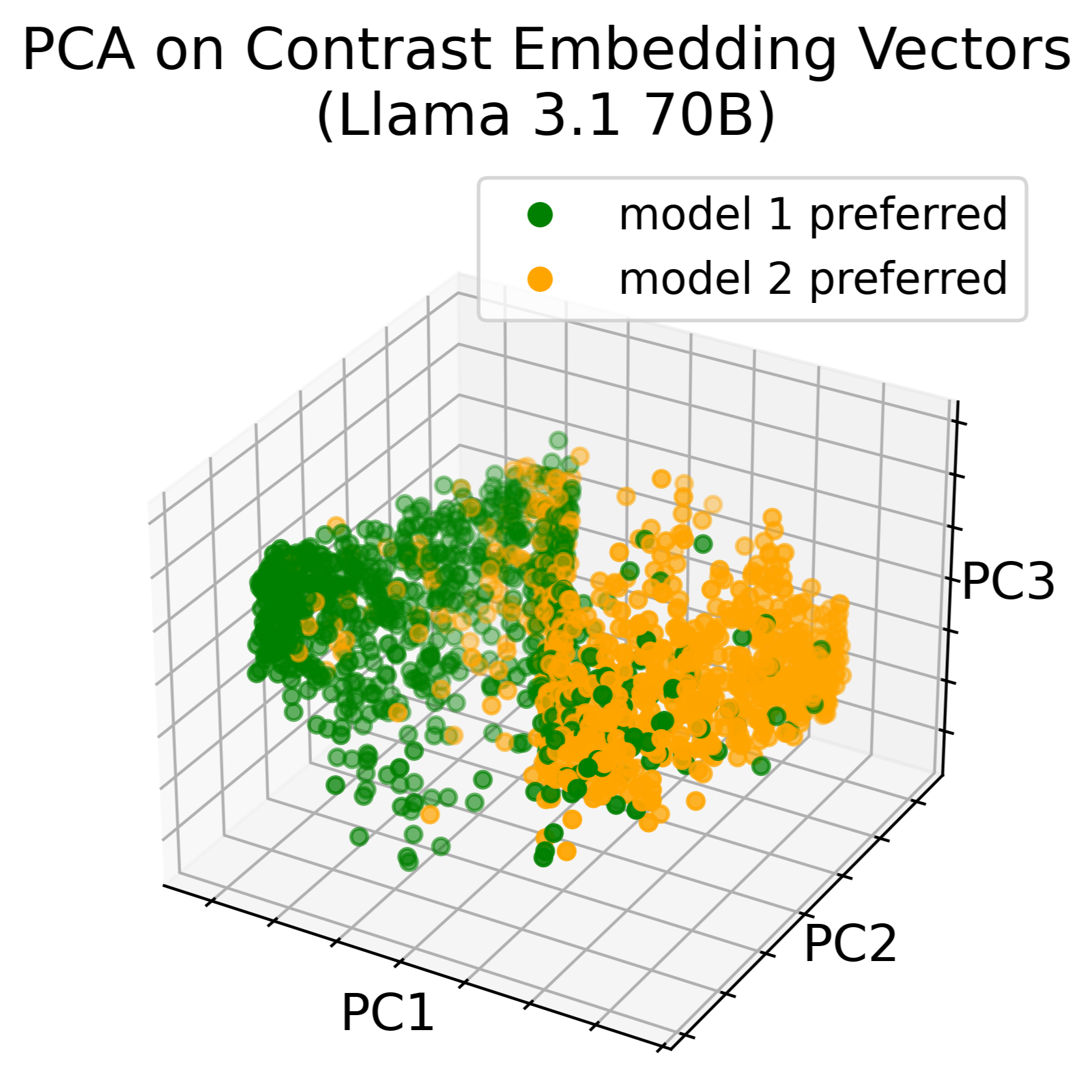} \hfill
    \includegraphics[width=0.4  \linewidth]{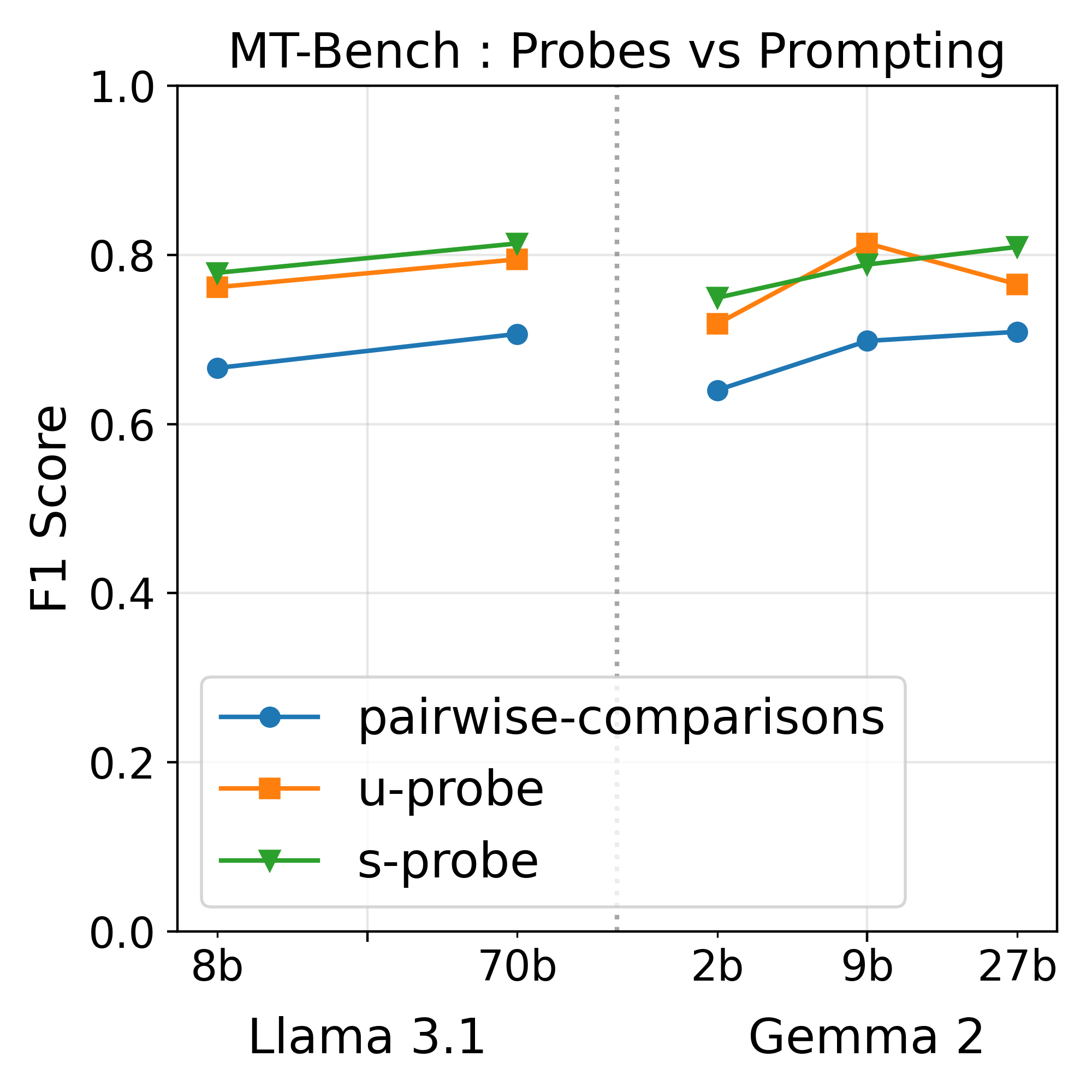}
    \caption {Our method exploits the empirical result that LLMs' internal features of ``belief'' or ``judgement'' are correlated with linear directions in their embedding spaces. For Llama 3.1 70B evaluated on the MT-Bench dataset, we find \textbf{the first principal component of the contrast pair differences of embedding vectors roughly classifies which model in a given example was preferred by a panel of human raters} \textit{(left)}. \textbf{Supervised or unsupervised classifying probes built on these embedding vectors are more aligned with human raters than prompting methods alone} \textit{(right)}, and this result holds across different model families (Gemma 2, Llama 3.1) at different sizes (from 2B to 70B parameters).}
    \label{fig:mtbench}
\end{figure*}

Chatbot Large Language Models (LLMs) are often trained using Reinforcement Learning with Human Feedback (RLHF) over preference datasets in order to increase honesty, helpfulness, and harmlessness \citep{christiano2017deep, stiennon2020learning, bai2022training}. This manifests as an increase in value/judgement alignment with humans, allowing for the use of such models as stand-in replacements for human raters on various tasks of evaluation \citep{zheng2023judging, shen2023large, zeng2023evaluating, stephan2024calculation, zhong-etal-2022-towards}. This approach, commonly known as \textit{LLM-as-a-Judge}, is particularly powerful for its fast and automatic nature.

In these tasks, LLM evaluators perform direct score-based assessment or pairwise comparisons, conventionally through generation-based prediction, where models are prompted to output numerical or Likert scale ratings, or comparative judgements. However, several factors limit the accuracy and efficiency of such approaches. Constrained-decoding for format control can introduce artifacts, and unintentional biases can be introduced by prompts. Overly verbose reasoning can obscure or misalign core judgements. More fundamentally, black-box approaches can lead to untrustworthy or factually incorrect generations, frequently stemming from biases learned during pretraining \citep{weidinger2021ethical, park2023ai, evans2021truthful, hendrycks2021unsolved}.

Previous work, such as \citet{liu2024aligning}, demonstrates that reformulating direct-scoring tasks as ranking problems based on pairwise preferences results in better alignment with human expert labellers. To extract even more accurate judgements than these generation-based approaches, we propose using classifying probes through pairwise comparisons. Empirical work suggests models' \textit{latent knowledge}, independent of the biases considered above, can be identified through the use of trained classifier heads on the activations of a given model. Such probes can be trained in a supervised \citep{alain2017understanding, marks2024the} or unsupervised \citep{burns2023discovering} fashion, and importantly, can be trained using \textit{contrast pairs}. This involves comparing the hidden state of a model when generating different possible answers, and observing salient \textit{contrastive features} in the change of hidden state, while controlling for irrelevant features \citep{burns2023discovering, rimsky-etal-2024-steering}. This leads both to better predictive performance and gains in efficiency.

We present the first thorough investigation of the performance of supervised and unsupervised probes for LLM-as-a-Judge tasks of pairwise preferences, comparing these methods to generation-based evaluation, with and without supervised finetuning (SFT). To summarise our main contributions:
\begin{itemize}[noitemsep]
    \itemsep 0em
    \item We introduce a way to extract human-aligned judgement from LLMs, by leveraging linear classifying probes and contrast pairs, in both a supervised and unsupervised setup.
    \item Through extensive experiments, we show classifying probes consistently outperform generation-based evaluation.
    \item We also show supervised probes considerably improve on the cost:performance ratio of SFT in realistic scenarios.
    \item We demonstrate these probes correlate with interpretable features of the underlying language model, generalising well to different domains and remaining more robust to distributional shifts than prompting.
\end{itemize}
Our results are consistent across four widely-used open-weights model families, at sizes ranging from 0.5B to 123B parameters, and six different datasets.
In light of our results, we encourage practitioners to make use of classifying probes for similar tasks for a more cost-efficient, robust, and performant solution.

\section{Background and Related Work} \label{sec:background}

\subsection{LLM-as-a-Judge} \label{subsec:llm_as_as_judge}

LLMs are increasingly employed as automatic, reference-free evaluators for assessing natural language tasks \citep{zhong-etal-2022-towards, chen2023exploring, wang-etal-2023-chatgpt, tan2024proxyqa}. Their applications span a wide range of domains, such as summarisation \citep{shen2023large}, instruction following \citep{zeng2023evaluating}, legal analysis \citep{deroy2024applicability}, reasoning \citep{stephan2024calculation}, and recommendation systems \citep{hou2024large}.
Despite their growing adoption, LLM-as-a-judge faces several challenges, including misalignment with human judgments \citep{chiang-lee-2023-large}, biases in various forms \citep{zheng2023judging, zhou-etal-2024-fairer}, inconsistencies in decision-making \citep{liu2024aligningb}, and limitations in reasoning capabilities \citep{stephan2024calculation}.

To address these issues, researchers have proposed several methods to enhance the reliability and accuracy of LLM-based judgments. G-Eval \citep{liu-etal-2023-g} refines scoring granularity using a logit-weighted average of score tokens. Pairwise comparison techniques have been introduced to improve alignment with human preferences, as demonstrated by \citet{liu2024aligning} and \citet{liusie-etal-2024-llm}. Other approaches advocate for generating Chain-of-Thought rationales prior to evaluation \citep{saha2025learning, wang2024selftaughtevaluators, ankner2024critique} or employing multi-agent debate frameworks \citep{chan2024chateval} or evaluator panels \citep{verga2024replacing} to enhance assessment robustness.

\subsection{Representation Probing} \label{subsec:probing}

Probing methods assess the extent to which language model representations encode specific knowledge. Typically, a probe is a supervised classifier \citep{conneau-etal-2018-cram, hupkes2018visualisation} trained to extract information from these representations \citep{christiano2021eliciting, belrose2023eliciting}. Linear probing \citep{alain2018understandingintermediatelayersusing}, which employs a linear classifier, is particularly valued for its efficiency and interpretability. Unsupervised variants have also been explored \citep{burns2023discovering, laurito-etal-2024-cluster}.

Probing has been widely applied to interpret model representations across various domains, including word embeddings \citep{levy2014linguistic}, sentiment \citep{maas2011learning}, factual knowledge \citep{marks2024the}, spatial and temporal understanding \citep{gurnee2024language}, and world models \citep{li2023emergent}. It has also been used to detect behavioural patterns such as outliers \citep{mallen2024eliciting}, inactive modules \citep{macdiarmid2024sleeperagentprobes}, and unfaithful generation \citep{azaria2023the, campbell2023localizing}.

In this work, we employ linear probing to extract evaluation judgments from an LLM-as-a-judge setup. Compared to inference-based or logits-based judgments, we show that linear probing improves both accuracy and efficiency.

\section{Methodology} \label{sec:methodology}

In order to identify an LLM's true ``judgement'' via classifying probes, we seek to identify binary features of belief or knowledge: a given text may or may not be consistent with the knowledge the LLM has learned during training, and we wish to identify a linear direction in activation space correlated with this property.

To identify such a direction, we make use of \textbf{contrast pairs} \citep{burns2023discovering}. We begin with a diverse set of binary statements or questions $S = \{s_i\}_{i=1}^N$. The contrast pairs are a dataset of prompts $X = \{(x_i^+, x_i^-)\}_{i=1}^N$ constructed by appending contrasting tokens to each $s_i$. Suppose for example that $s_i =$ ``The capital of France is Paris.'' A contrast pair for factual accuracy on $s_i$ would have $x_i^+ =$ ``The capital of France is Paris. This statement is true'' and $x_i^- =$ ``The capital of France is Paris. This statement is false''. 

Both prompts are then used as inputs to an LLM, and the embedding vectors $\phi(x_i^+)$ and $\phi(x_i^-)$ of the differing contrasting tokens are harvested at a layer $l$.

We assume both $\phi(x_i^+)$ and $\phi(x_i^-)$ can be decomposed into several features, most of which are shared (since both are derived from the statement $s_i$). We also assume we can approximate both as a linear combination of said features:
\begin{align*}
    \phi(x_i^+) &= \sum_{i=1}^{n}\mathcal{F}^{shared}_{i} + \sum_{j=1}^{m}\mathcal{F}^{+}_{i} + \epsilon^+, \\
    \phi(x_i^-) &= \sum_{i=1}^{n}\mathcal{F}^{shared}_{i} + \sum_{j=1}^{k}\mathcal{F}^{-}_{i} + \epsilon^-,
\end{align*}
with all $\mathcal{F}^{shared}$ common to both embeddings, $\mathcal{F}^{+/-}$ unique to each element of the contrast pair and remaining information $\epsilon^{+/-}$.

Consider the contrast pair difference $\phi(x_i^+) - \phi(x_i^-)$, removing the effect of all $\mathcal{F}^{shared}$ and leaving only \textbf{contrastive features}. Two immediately obvious contrastive features are:

\begin{itemize}[noitemsep]
    \itemsep 0em
    \item $\Delta_{syntax} := \mathcal{F}^{True} - \mathcal{F}^{False},$ the syntactical difference in the prompts $x^+$ and $x^-$.
    \item $\Delta_{knowledge} := \mathcal{F}^{\top} - \mathcal{F}^{\bot},$ the logical difference between both prompts: one is consistent with the model's internal knowledge while the other is not. This can be thought of as the model's ``belief'' in a sense.
\end{itemize}

Given a dataset of contrast pair differences $D = \{\phi(x_i^+) - \phi(x_i^-)\}_{i=1}^N$, a centering step can be performed to remove $\Delta_{syntax}$ before taking this difference:
\begin{align*}
    \tilde{\phi}(x_i^+) &:= \phi(x_i^+) - \mu^+, \\
    \tilde{\phi}(x_i^-) &:= \phi(x_i^-) - \mu^-,
\end{align*}
where $\mu^+$ and $\mu^-$ are the mean embedding vectors of $\{\phi(x_i^+)\}$ and $\{\phi(x_i^-)\}$ respectively. We claim $\Delta_{knowledge}$ will, in most cases, be the \textit{most salient} contrastive feature of the new dataset $\tilde{D} = \{\tilde{\phi}(x_i^+) - \tilde{\phi}(x_i^-)\}_{i=1}^N$.

Given ground truth labels for each pairwise comparison, we can model the probability with a classifier:
\begin{align*}
    \mathbb{P}(x^+\text{ true}) = \sigma(\mathbf{w}^T(\tilde{\phi}(x_i^+) - \tilde{\phi}(x_i^-))).
\end{align*}
The supervised probes we train in Section~\ref{sec:experiments} are described by the above classifier, the parameters of which we fit using logistic regression.

Additionally, were the above claim of salience of $\Delta_{knowledge}$ to be true, it would be identifiable as the top principal component of our dataset, thereby allowing us to obtain a probing classifier without the need for ground-truth labels. Indeed, this approach is commonly used as an unsupervised probing technique for similar tasks \citep{burns2023discovering, laurito-etal-2024-cluster}, and we explore its use for LLM-as-a-Judge pairwise comparisons here too.

\section{Experimental Setup} \label{sec:experiments}

\begin{figure*}[!ht]
    \centering
    \includegraphics[width=0.7\textwidth]{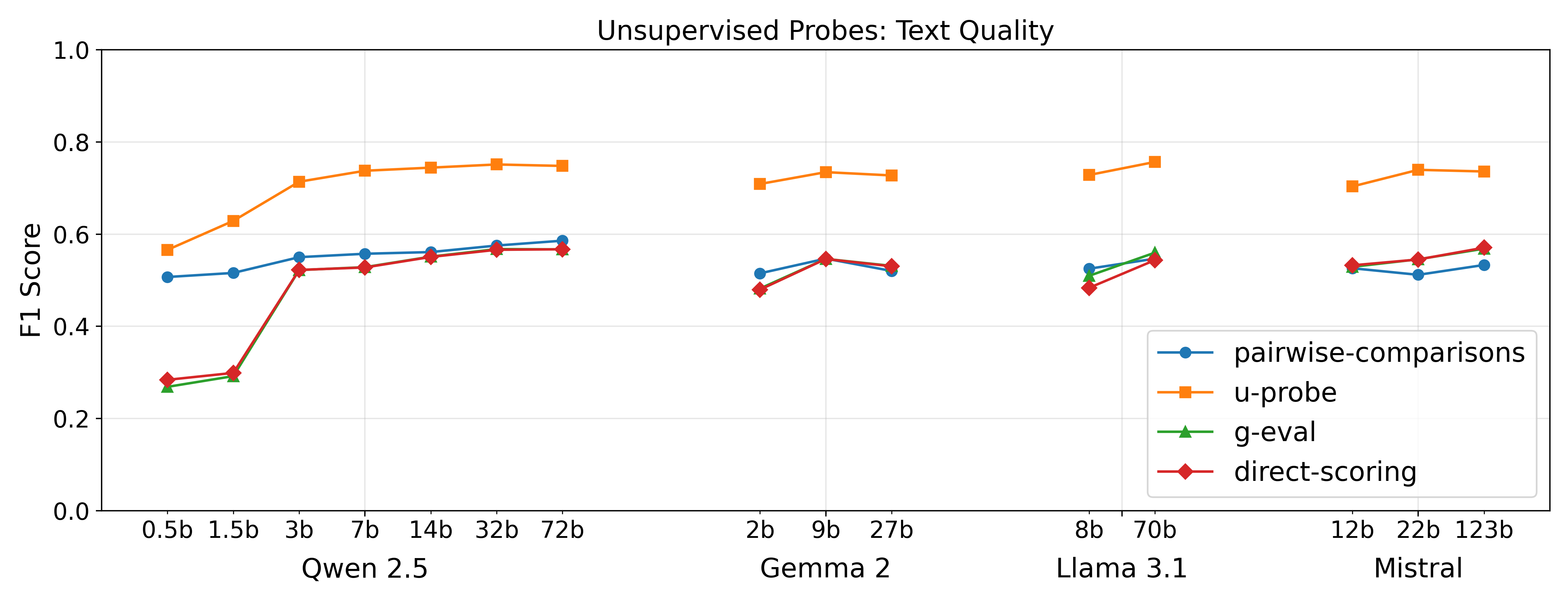}
    \vspace{0.5em}
    \includegraphics[width=0.7\textwidth]{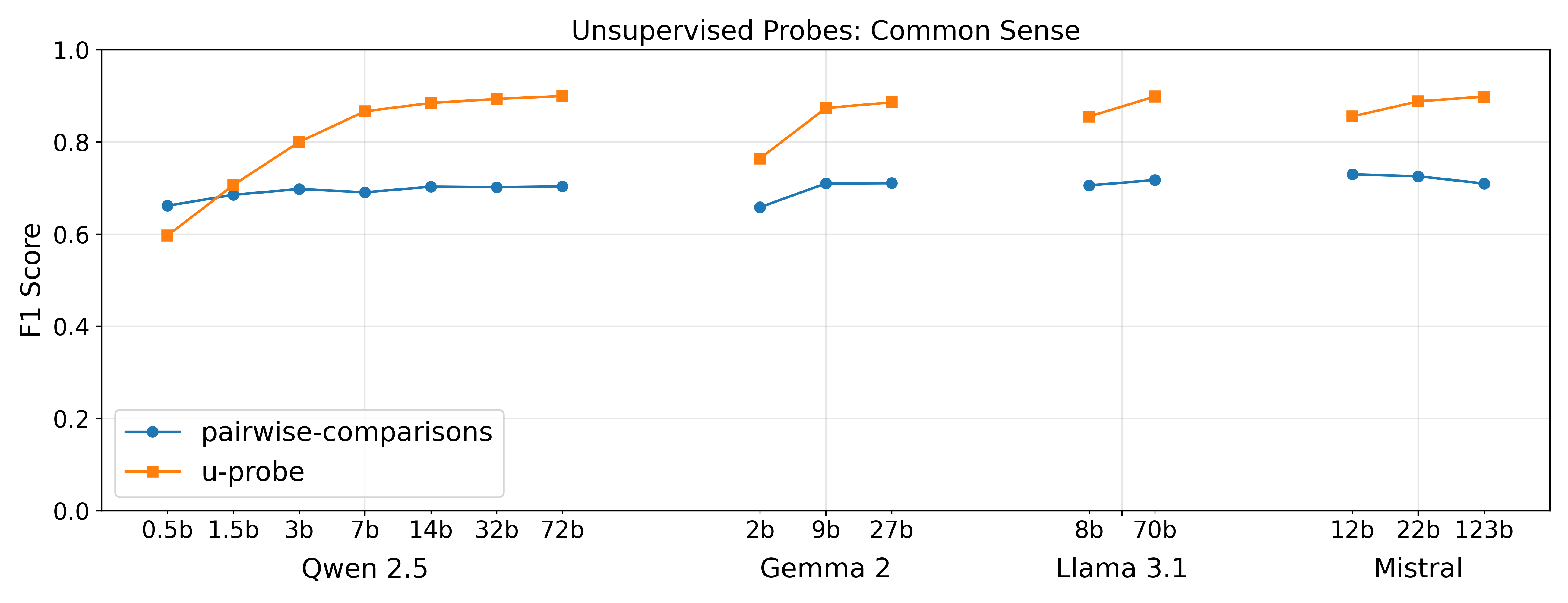}
    \caption{\textbf{Unsupervised probes, in all but one test case, outperform generation-based methods like direct-scoring and pairwise comparisons.} Interestingly, within a given model family, unsupervised probing performance with a small model almost always outperforms prompting performance with much larger models. This highlights two related key findings: (1) the use of relatively large LLMs for LLM-as-a-Judge tasks may be unnecessarily computationally wasteful and (2) there may be significant capability ``left on the table'' with smaller LLMs for such tasks.}
    \label{fig:unsupervised_probes}
\end{figure*}

\subsection{Datasets} \label{subsec:datasets}

We investigate the performance of classifying probes for LLM-as-a-Judge tasks through the use of several datasets spanning different problem domains.

\textbf{LLM Chat Preferences} MT-Bench \citep{NEURIPS2023_91f18a12} is a multi-turn question set of pairwise comparisons of chatbot LLM interactions. A subset of these comparisons have been performed by several human labellers, and we use these as ground-truth to measure performance against.

\textbf{Text Quality} The NEWSROOM \citep{grusky-etal-2018-newsroom}, SummEval \citep{fabbri2020summeval}, and HANNA \citep{chhun2024do} datasets all concern the evaluation of text in terms of high-level concepts. News articles with summaries (former two) and story prompts with short stories (last) are evaluated by human labellers on several high-level features such as ``coherence'' or ``surprise''. Note these are directly scored on a Likert scale; when necessary we convert this task to one of pairwise comparisons, following \citet{liu2024aligning}.

\textbf{Common Sense Reasoning} The ROCStories dataset \citep{mostafazadeh-etal-2016-corpus} consists of short story prompts provided with two potential endings. One ending is always more consistent with the story prompt, allowing for a pairwise comparison task. We additionally make use of the MCTACO \citep{zhou-etal-2019-going} and CaTeRS \citep{mostafazadeh-etal-2016-caters} datasets which similarly test common sense reasoning in the context of causal/temporal understanding.

\subsubsection{Prompts} \label{subsubsec:prompts}

Full prompt templates for all datasets can be found in Appendix~\ref{app:prompts} or in our code repository to be released. However, the general prompt template used in all pairwise comparison experiments is shown below:
\begin{verbatim}    
Consider the following two <items>:
Choice 1: <item 1>
Choice 2: <item 2>
Which is more <task>? 
Answers must be a single choice.
\end{verbatim}
When harvesting contrast pairs, we prime the model with the following message:
\begin{verbatim}
Between Choice 1 and Choice 2, the more 
<task> <item> is Choice <contrast_token>
\end{verbatim}

\subsection{Models} \label{subsec:models}

Our results robustly generalise between different LLM model families. We demonstrate this in Section~\ref{sec:results}, where we conduct scaling analyses only within model families, as idiosyncratic differences between them lead to different patterns of performance and scaling. Specifically, we consider:
\begin{itemize}[noitemsep]
    \itemsep 0em
    \item the two smaller (8B, 70B) Llama 3.1 models \citep{grattafiori2024llama3herdmodels}.
    \item the Gemma 2 (2B, 9B, 27B) family of models \citep{gemmateam2024gemma2improvingopen}.
    \item the Qwen 2.5 (0.5B, 1.5B, 3B, 7B, 14B, 32B, 72B) family of models \citep{qwen2025qwen25technicalreport}.
    \item Mistral Nemo (12B), Small (22B), and Large (123B) \citep{mistral_nemo, mistral_small, mistral_large}.
\end{itemize}
All our experiments are conducted through question-answering, and due to this \textbf{all} models we use have undergone a post-training pipeline of (usually) instruction-tuning and some form of preference learning such as reinforcement learning from human feedback \citep{christiano2017deep}.

\subsection{Baselines} \label{subsec:baselines}

For the text quality datasets mentioned in Section~\ref{subsec:datasets} we report baseline results of generation-based prompting a given model to evaluate text on the original Likert scale e.g., on a scale of 1 to 5, referring to this as \textit{direct-scoring}. We additionally report a recent improvement to this approach, G-Eval \citep{liu-etal-2023-g}, which re-weights predictions using the model's own predicted probabilities for each possible answer choice.

When re-framing the above tasks as pairwise comparisons, and with all other datasets, we report prompting performance for comparisons. To address positional bias, we marginalise over position and take an average of the model's predicted probabilities. Note: this assumes a consistent positional bias, and requires us to run each question through the model twice (with the two possible answer choices swapped).

\subsection{Training Setup} \label{subsec:training_setup}

Generation-based prediction is performed by examining model predicted probabilities for possible answer choices e.g., for pairwise comparisons, we compare the probabilities for the tokens ``1'' and ``2''. For activation harvesting, unless otherwise stated, we take the embedding vector of the final token (that is, the contrasting token) of a given prompt, after the last decoder block and before the final normalisation layer\footnote{The choice of layer is further investigated in Appendix~\ref{app:layer}.}.

Both supervised and unsupervised probes are fit and tested on random distinct halves of a given dataset.

\section{Results} \label{sec:results}

\begin{figure*}[!ht]
    \centering
    \includegraphics[width=0.7\textwidth]{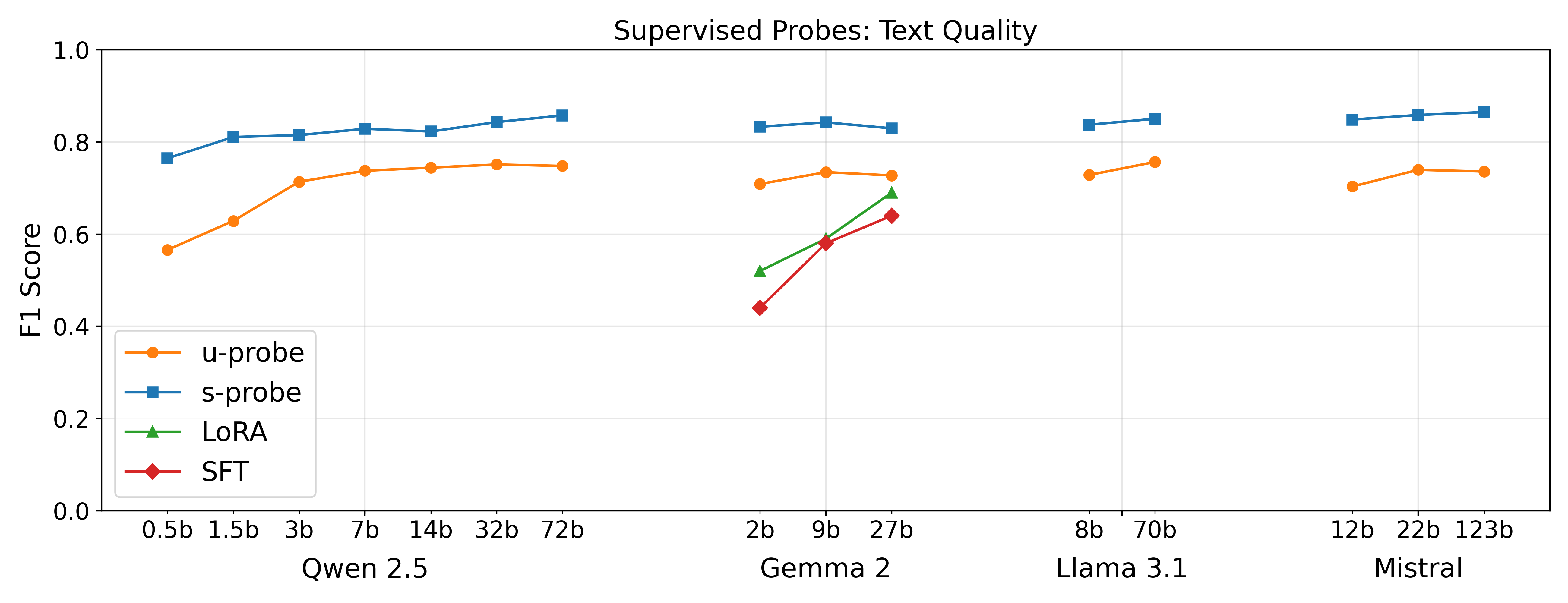}
    \vspace{1em}
    \includegraphics[width=0.7\textwidth]{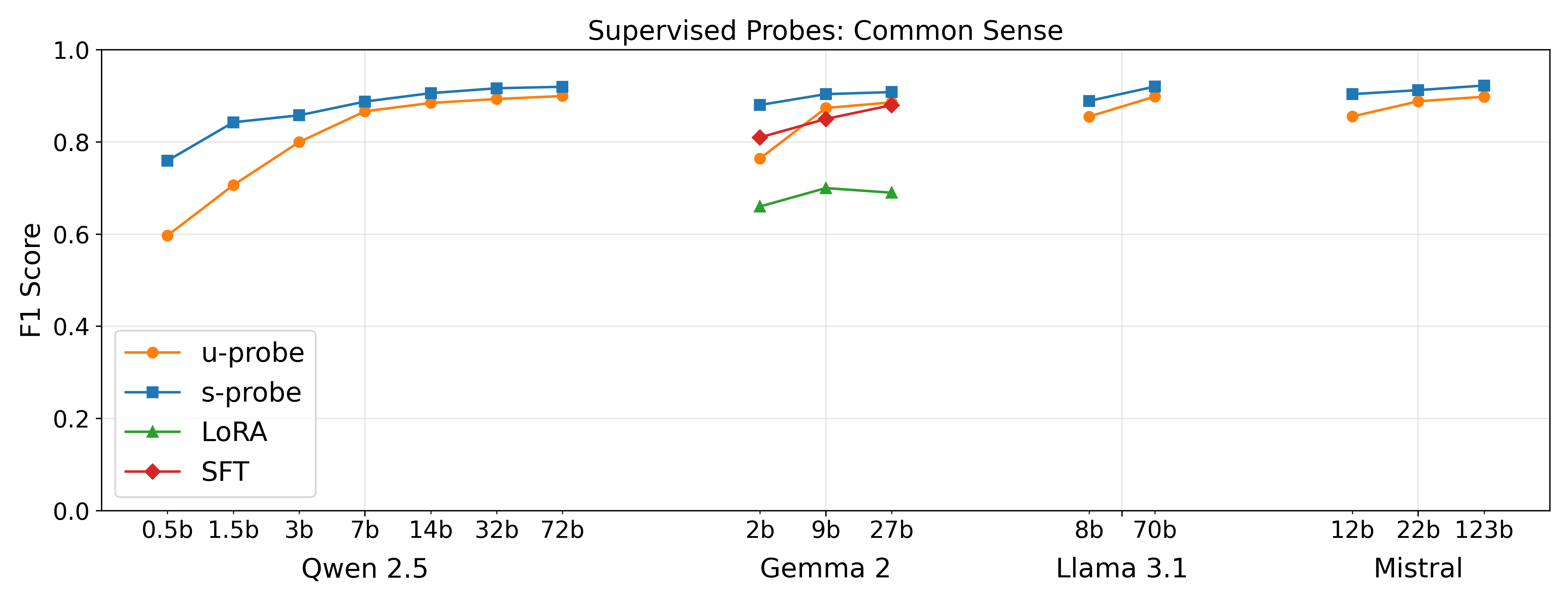}
    \caption{\textbf{Supervised probes, in all cases, allow for a further improvement in alignment with human raters over unsupervised probes.} We also test parameter-efficient and full finetuning of models in the Gemma 2 family, finding that supervised probes still outperform finetuned generation-based evaluators.}
    \label{fig:supervised_probes}
\end{figure*}

\begin{figure*}[!ht]
    \centering
    \includegraphics[width=0.7\textwidth]{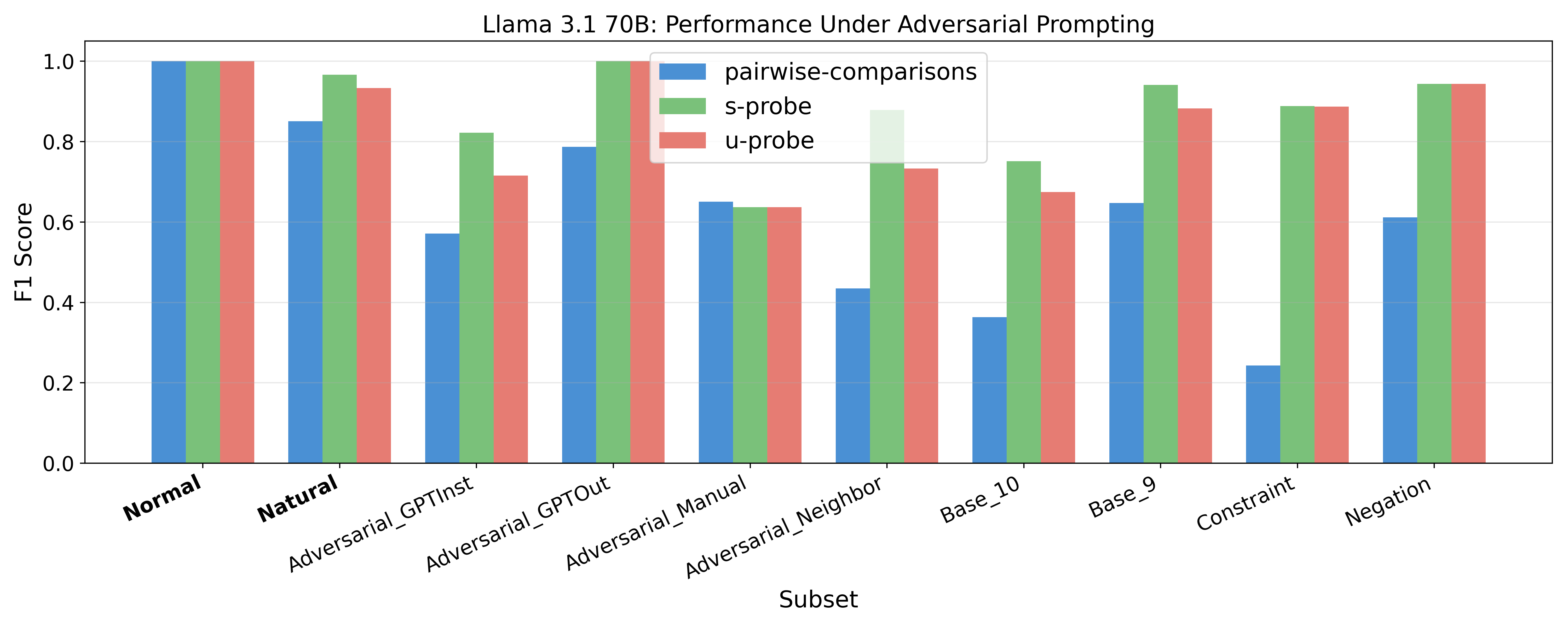}
    \vspace{0.5em}
    \caption{Performance of classifying probes and generation-based prompting for Llama 3.1 70B on the LLMBar dataset. All three methods suffer under adversarial prompting (non-bold subsets), however, \textbf{both probing approaches remain significantly more robust than prompting.}}
    \label{fig:ablation_llama70b}
\end{figure*}

We first use the MT-Bench dataset to assess the ability of different LLMs to compare two model-generated answers to a user-question in a chatbot interaction: a common task in LLM post-training. A panel of human judges reached 80\% agreement on this dataset \citep{NEURIPS2023_91f18a12}, and we obtain ground-truth labels by taking the majority-vote of this panel. Both supervised and unsupervised probes perform similarly at aligning with this ground-truth, achieving F1 scores of roughly 0.8, as can be seen in Figure~\ref{fig:mtbench}. Importantly, we find classifying probes outperform prompting methods, while maintaining the same inference cost of two forward passes per example. This motivates our deeper investigation into the potential of classifying probes for similar tasks, which we present now.

\subsection{Experiment 1: Unsupervised Probes} \label{subsec:unsupervised_probes}

We analyse the performance of the PCA-based unsupervised probing method described in Section~\ref{sec:methodology} through the three \textbf{text quality} datasets NEWSROOM, SummEval, and Hanna, and the three \textbf{common sense reasoning} datasets ROCStories, MCTACO, and CaTeRS. A baseline for our probes should also be unsupervised; we compare against zero-shot prompting on all datasets, calibrating model predictions by running each input example twice, swapping the order of examples in a given pairwise comparison, allowing us to marginalise over answer position to account for order effects. In the case of the text quality datasets we can also report direct scoring on the original Likert scale of 1-5, and a G-Eval-based \citep{liu-etal-2023-g} correction of this approach.

\paragraph{Unsupervised Probes Outperform Calibrated Prompting Methods}
We find for all six datasets and four model families, aside from a single test case (Qwen 2.5 0.5B), the use of unsupervised probes allows for significantly higher alignment with human judgement (Figure~\ref{fig:unsupervised_probes}). We see this as evidence of a capability-gap between models' abilities measured through the flexibility and capacity of their latent spaces and their abilities measured through standard prompting approaches. Such a gap may narrow over time with the release of newer models with higher instruction-following capabilities, but it remains sizable for now. We therefore advocate the use of unsupervised probes for pairwise comparison tasks in which labels are sparse/absent over prompting methods alone.

These results, further broken down into each of the six constituent datasets aggregated in Figure~\ref{fig:unsupervised_probes}, can be found in Appendix~\ref{app:full_probes}.

\subsection{Experiment 2: Supervised Probes} \label{subsec:supervised_probes}

For many LLM-as-a-Judge tasks, it may be feasible to obtain a (small) number of labelled examples to guide the decision-making process of an LLM evaluator. In such cases, a supervised probe can be trained by replacing the PCA step above with a standard supervised classifier, as described in Section~\ref{sec:methodology}. For the same six datasets evaluated above, we train supervised probes on 5000 examples and examine their performance against corresponding unsupervised probes on the remaining held-out examples. 

\paragraph{Supervised Probes Outperform Unsupervised Probes And Can Outperform Finetuning}
We find, as shown in Figure~\ref{fig:supervised_probes}, that supervised probes often allow for a \textit{further} increase in alignment with human raters. For particularly sensitive tasks in which supervised approaches are feasible, practitioners may opt to finetune a given model to improve its performance. We also find supervised probes are a competitive alternative to such an approach, as shown for the Gemma 2 model family in Figure~\ref{fig:supervised_probes}. For both the text quality and common sense reasoning tasks, supervised probes outperform LoRA \citep{hu2022lora} and even full finetuning with the same number of training examples, at all model sizes
\footnote{Details on the finetuning process performed are provided in Appendix~\ref{app:finetuning}.}. 

The results in Figure~\ref{fig:supervised_probes} may shed some light on the difficulty of the text quality and common sense reasoning tasks set up in our experiments. Note for the former, finetuned models actually perform relatively poorly, with a large capability gap against both unsupervised and supervised probes. In the common sense reasoning task, finetuning is much more competitive. We hypothesize this is due to the subjective vs objective nature of the two tasks. The evaluation of text on abstract features such as ``coherence'' and ``empathy'' (as is carried out in the NEWSROOM, SummEval, and HANNA datasets) is likely highly subjective, while common sense reasoning can be considered a much more objective task. This makes the latter much easier to learn during pretraining, and further improve on during finetuning. This is further reflected in the larger improvement in supervised over unsupervised probes for the text quality task: human-generated labels allow the probe-fitting process to efficiently align with raters. Conversely, finetuning approaches likely require many more labels to converge to this same distribution, with orders of magnitude more parameters requiring tuning over the logistic regression classifiers we train.

This suggests a key advantage of probing approaches constructed through contrast pairs: the salience of the desired knowledge or belief is increased, facilitating easier learning of the task and reductions in computational cost compared to finetuning. We expect, in the limiting case of labelled data, finetuning approaches will overtake probe performance due to their higher flexibility. However, for many realistic applications, labelled data can be unreasonably expensive to obtain.

These results, broken down into each of the six constituent datasets aggregated in Figure~\ref{fig:supervised_probes}, can be found in Appendix~\ref{app:full_probes}.

\subsection{Experiment 3: Probe Generalisation} \label{subsec:generalisation}

In addition to offering advantages in both computational cost and performance for LLM-as-a-Judge tasks, classifying probes yield key interpretability insights into LLMs in general. We find evidence they correlate with \textit{general} features of belief or judgement used by a given model.

\begin{table*}[t]
    \centering
    \resizebox{\linewidth}{!}{
        \begin{tabular}{|l | l | c c c c c c|}
        \cline{3-8}
        \multicolumn{2}{c|}{F1-Score} & \multicolumn{6}{c|}{Probe} \\
        \cline{3-8}
        \multicolumn{2}{c|}{Supervised/Unsupervised} & \multicolumn{1}{>{\columncolor{customgray}}c}{NEWSROOM} & \multicolumn{1}{>{\columncolor{customgray}}c}{SummEval} & \multicolumn{1}{>{\columncolor{customgray}}c}{HANNA} & \multicolumn{1}{>{\columncolor{customgray}}c}{ROCStories} & \multicolumn{1}{>{\columncolor{customgray}}c}{MCTACO} & \multicolumn{1}{>{\columncolor{customgray}}c|}{CaTeRS} \\
        \hline
        \multirow{6}{*}{\rotatebox[origin=c]{90}{Evaluation}} & \multicolumn{1}{>{\columncolor{customgray}}l|}{NEWSROOM} & - & 0.63/\textcolor{teal}{0.77} & 0.65/\textcolor{teal}{0.77} & \textcolor{teal}{0.78}/0.77 & \textcolor{teal}{0.78}/0.77 & 0.64/\textcolor{teal}{0.77} \\
        & \multicolumn{1}{>{\columncolor{customgray}}l|}{SummEval} & 0.66/\textcolor{teal}{0.76} & - & 0.63/\textcolor{teal}{0.76} & \textcolor{teal}{0.77}/0.76 & \textcolor{teal}{0.77}/0.76 & 0.58/\textcolor{teal}{0.76} \\
        & \multicolumn{1}{>{\columncolor{customgray}}l|}{HANNA} & 0.67/\textcolor{teal}{0.71} & 0.62/\textcolor{teal}{0.71} & - & 0.71/0.71 & 0.71/0.71 & 0.59/\textcolor{teal}{0.70} \\
        & \multicolumn{1}{>{\columncolor{customgray}}l|}{ROCStories} & 0.87/\textcolor{teal}{0.99} & 0.78/\textcolor{teal}{0.99} & 0.79/\textcolor{teal}{0.98} & - & 0.99/0.99 & 0.71/\textcolor{teal}{0.99} \\
        & \multicolumn{1}{>{\columncolor{customgray}}l|}{MCTACO} & 0.79/\textcolor{teal}{0.95} & 0.67/\textcolor{teal}{0.95} & 0.70/\textcolor{teal}{0.95} & 0.96/\textcolor{teal}{0.95} & - & 0.74/\textcolor{teal}{0.95} \\
        & \multicolumn{1}{>{\columncolor{customgray}}l|}{CaTeRS} & 0.75/\textcolor{teal}{0.78} & 0.62/\textcolor{teal}{0.78} & 0.67/\textcolor{teal}{0.78} & 0.76/\textcolor{teal}{0.78} & 0.77/\textcolor{teal}{0.78} & - \\
        \hline
        \end{tabular}
    }
    \vspace{0.5em}
    \caption{\textbf{Generalisation of classifying probes for Llama 3.1 70B.} We train both supervised \textit{(left)} and unsupervised \textit{(right)} probes on examples from a given dataset \textit{(columns)}, testing them on all five other datasets \textit{(rows)} through F1-score. The higher scoring probe of a given supervised/unsupervised pair is coloured. We find both sets of probes generalise relatively well, and unsupervised probes in particular generalise very well on several occasions.}
    \label{tab:generalisation}
\end{table*}

\begin{figure}[h]
  \includegraphics[width=\columnwidth]{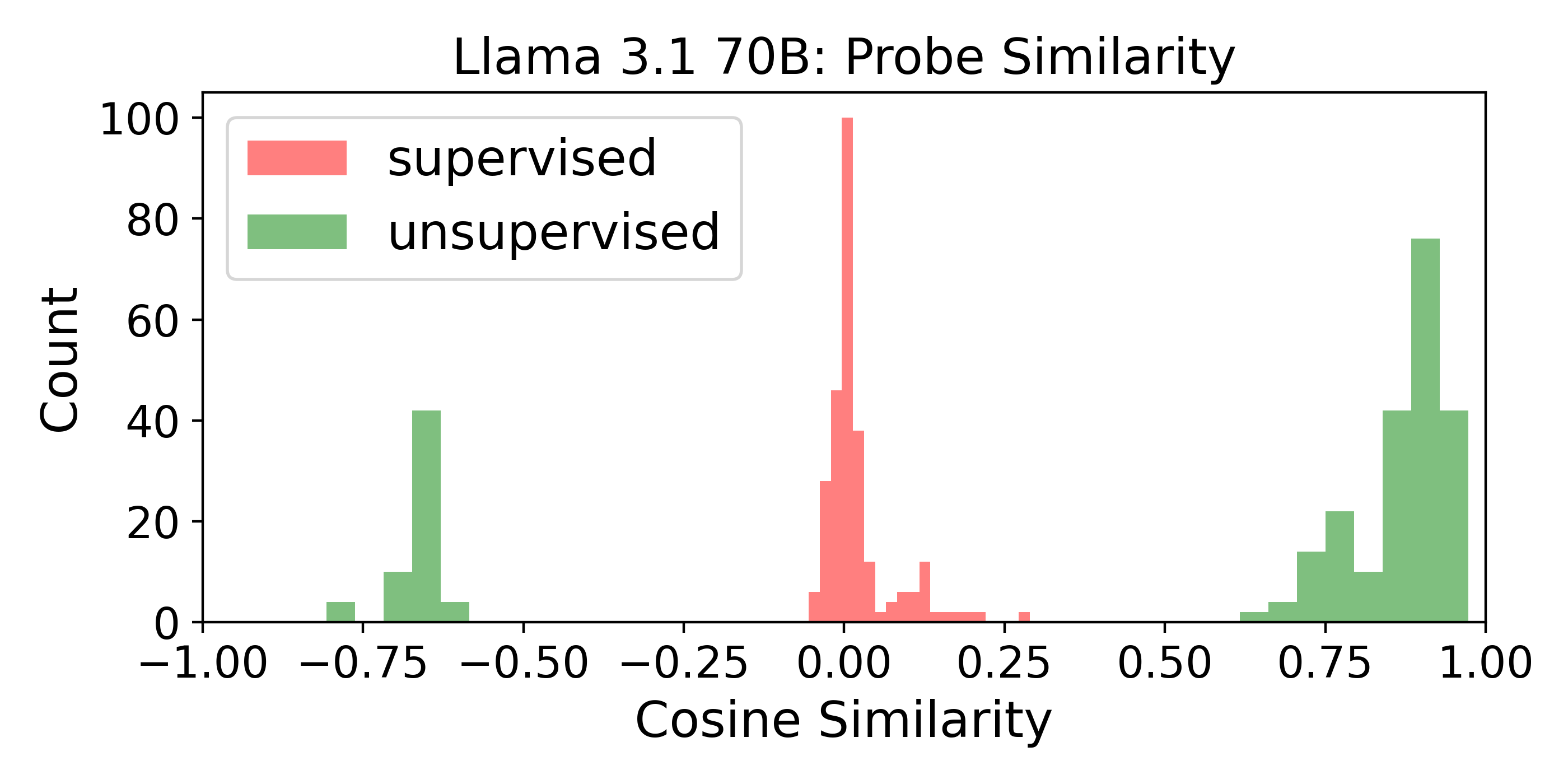}
  \caption[width=\columnwidth]{Taking the example of Llama 3.1 70B, we find most supervised probes are dissimilar while most unsupervised probes are similar (up to sign), regardless of the varying tasks in each of the six datasets considered.}
  \label{fig:cosine}
\end{figure}

This evidence, summarised in Table~\ref{tab:generalisation}, comes from an experiment into the generalisation of probes under significant distributional changes. Specifically, we train both supervised and unsupervised probes using contrast pair differences of activations from Llama 3.1 70B on each of the six datasets examined in the above experiments, and test each on the remaining five other datasets. 

\paragraph{Classifying Probes Identify Generalising Features Of Belief Or Judgement}
Generally, these probes achieve high F1 scores, even when trained and tested on very different tasks. Unsupervised probes in particular generalise extremely well in several cases, achieving F1 scores at or above 0.95. We hypothesize the contrast pair setup allows probes to focus on \textit{task-independent} features of judgement, relying on models' already vast knowledge and human alignment due to pretraining to infer the correct choice. Supervised probes can leverage information about a specific task, but this ultimately pushes the probe direction away from these task-independent features, leading to slightly worse generalisation than their unsupervised variants.

The better generalisation of unsupervised probes is supported by Figure~\ref{fig:cosine}, in which the cosine similarity between all probe directions across all datasets is plotted. Up to a sign flip, unsupervised probes are highly similar, with most having magnitude similarity above 0.7. Meanwhile, the distribution for supervised probes is narrowly centred around zero. It is possible that supervised probe similarity could be increased by training on more/diverse data, but it is striking that unsupervised probes trained on relatively different domains identify similar features. It is unclear however whether these features are causally relevant during the forward pass, to represent belief and judgement, but we investigate this further in Appendix~\ref{app:causal}.

\subsection{Experiment 4: Ablation Study} \label{subsec:ablations}

As a final test of classifying probes, we perform an ablation study of performance under different types of adversarial prompting strategies. To do so, we make use of the LLMBar dataset \citep{zeng2024llmbar} for evaluating instruction-following capabilities. This dataset is split into several subsets, all of which, other than the \textit{Normal} and \textit{Natural} subsets, are specifically designed to induce incorrect answers from LLM evaluators.

\paragraph{Classifying Probes Are More Robust to Domain Shifts Than Prompting}
We train probes on the \textit{Normal} and \textit{Natural} subsets only, testing them on all other subsets, and comparing with generation-based prompting as before. Figure~\ref{fig:ablation_llama70b} shows, for Llama 3.1 70b, how all methods suffer a performance drop under adversarial prompts. However, we note for all but one subset, this drop is significantly less severe for probing approaches over prompting; note in particular the results on the \textit{Constraint} subset for example. This finding holds across different model sizes and families - our replications of this experiment on other models can be found in Appendix~\ref{app:ablation}. This complements our results on probe generalisation: the relative robustness of classifying probes likely aids their ability to generalise to different domains.

\section{Conclusion}

We explore the use of linear classifying probes, both supervised and unsupervised, to perform pairwise comparisons in several standard LLM-as-a-Judge tasks. Our approach of using contrast pair differences to increase the salience of relevant ``belief'' or ``judgement'' features proves to be greatly effective; unsupervised probes consistently outperform calibrated generation-based evaluators across several open-weights LLM families and model sizes, without a significant increase in computational cost. In realistic scenarios with limited but available ground-truth labels, we also find supervised probes outperform unsupervised methods and can even outperform finetuning of the same model. These probes generalise well to different domains, and are more robust to (adversarial) distributional shifts than prompting approaches. Our experiments constitute the first comprehensive assessment of both supervised and unsupervised probes for LLM-as-a-Judge tasks against generation-based approaches, with and without finetuning. They suggest for practical applications, classifying probes are a cost-efficient, robust, and powerful solution.

\section*{Limitations}

Our experiments in Section~\ref{subsec:supervised_probes} find supervised probes outperform finetuned (both LoRA \citep{hu2022lora} and full) generation-based evaluators given the same training data. It could be interesting to investigate how and when probe performance saturates, and relatedly whether finetuning approaches outperform probes in the limit of data availability. While this is outside the scope of our study, future work establishing the threshold at which this may (or may not) take place would better inform developers of best practices. 

Additionally, we focus the scope of LLM-as-a-Judge tasks covered in this work to those of pairwise preferences, as this setup lends itself well to the use of binary classifying probes. We would be excited to see future work exploring the use of latent knowledge in direct-scoring tasks, where texts are rated on a numerical or Likert scale. This could be achieved through one-vs-rest or multi-class probes for example.

Finally, there remain additional challenges to overcome with probing methods in general. Red-teaming studies and analyses \citep{farquhar2023challengesunsupervisedllmknowledge, laurito-etal-2024-cluster} find that prompts which can induce a language model into simulating a different \textit{quality} of knowledge e.g., \textit{``You are a smart professor...''}, can significantly affect probe performance. Addressing this challenge proves to be particularly difficult for the research community, as it requires a much better understanding of knowledge representation within LLMs. For now, this presents a fundamental limitation of probing and other similar white-box approaches.

\section*{Responsible NLP Statement}

We follow strict compliance with all dataset and model licenses relevant in this work. AI assistants were used in the process of writing experimental code only.

\section*{Acknowledgments}

SM is grateful for support received by the UKRI Centre for Doctoral Training in Application of Artificial Intelligence to the study of Environmental Risks [EP/S022961/1].

\bibliography{custom}
\clearpage
\newpage
\appendix

\section{On The Target Layer For Activation Harvesting} \label{app:layer}

All classifying probes examined in Section~\ref{sec:results} are trained following the same process. One key step, as explained in Section~\ref{sec:methodology}, involves the harvesting of activations during the forward pass. In the contrast pair setup, we do so on the token position of contrasting tokens at the last layer of a given model. Here, we briefly explore how necessary this choice is.

\begin{figure*}[h]
    \centering
    \includegraphics[width=0.7\textwidth]{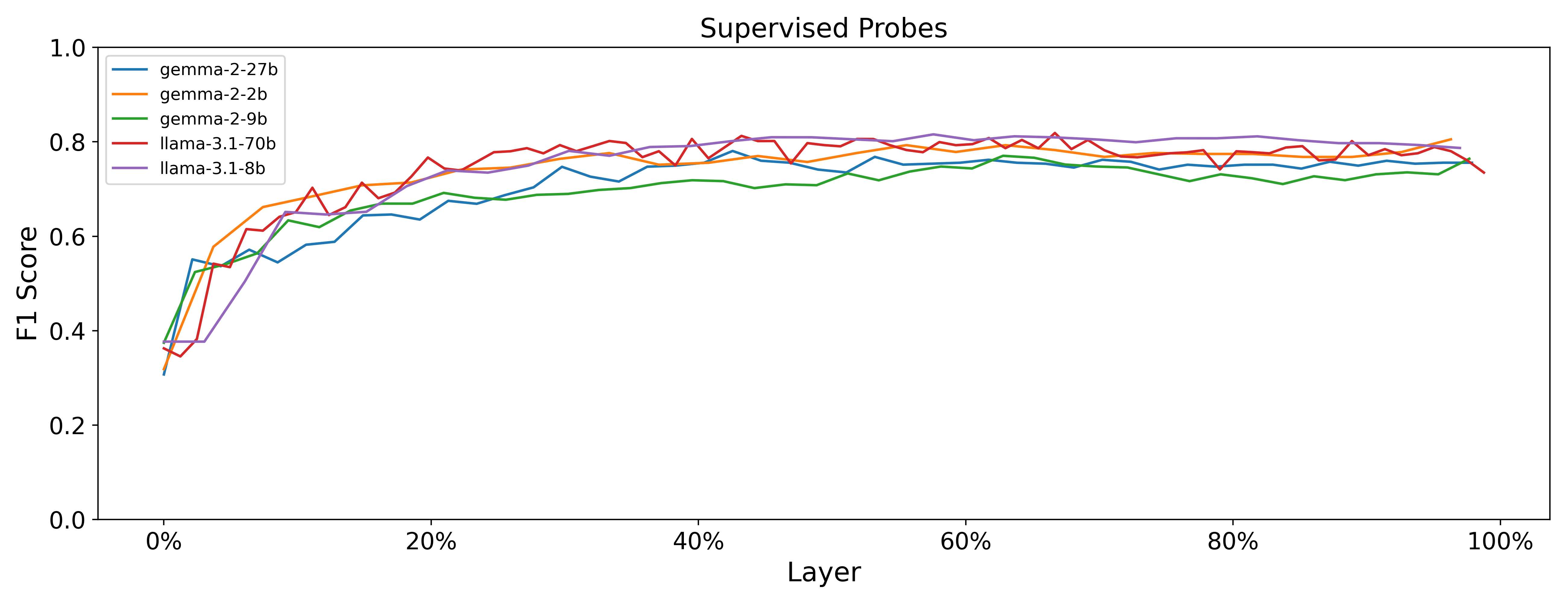}
    \caption{Performance of supervised probes trained on all layers of a given language model and evaluated on the MT-Bench dataset. We see a relatively \textbf{smooth increase in probe performance through the forward pass.}}
    \label{fig:supervised_layers}
\end{figure*}

\begin{figure*}[]
    \centering
    \includegraphics[width=0.7\textwidth]{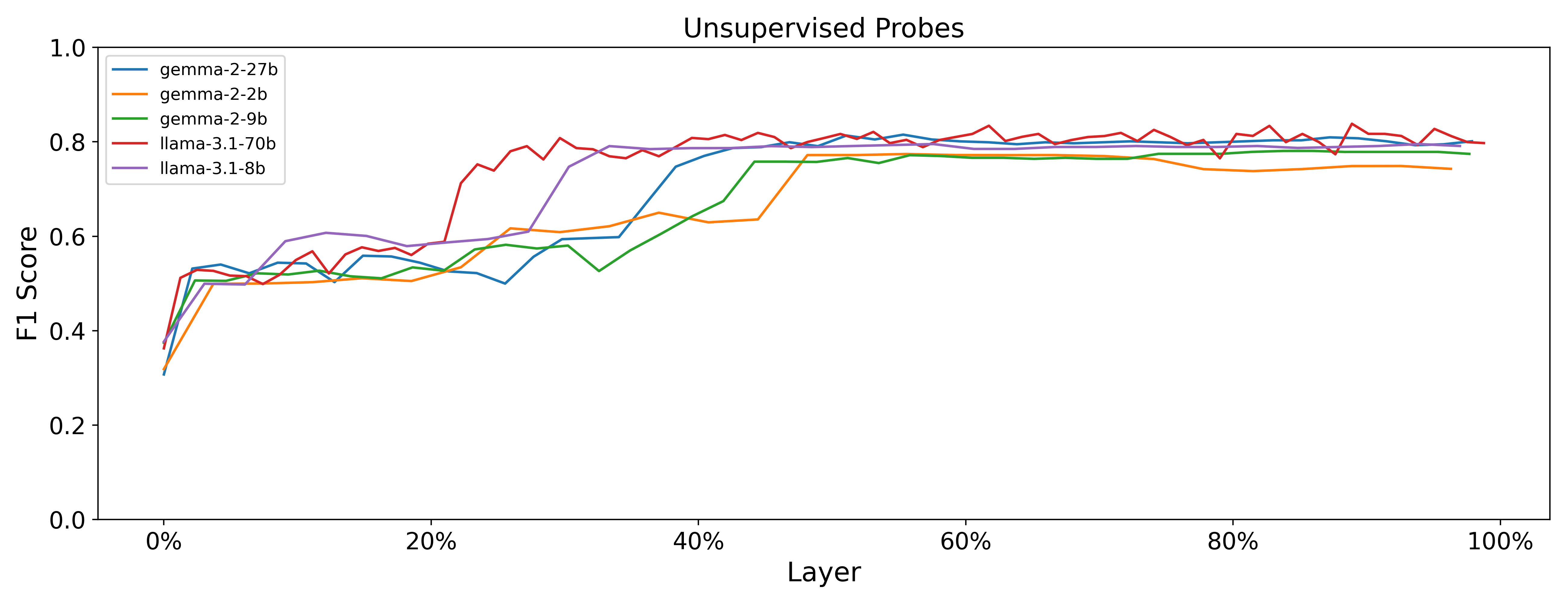}
    \caption{Performance of unsupervised probes trained on all layers of a given language model and evaluated on the MT-Bench dataset. In contrast with Figure~\ref{fig:supervised_layers}, we see a \textbf{discontinuous jump in performance of unsupervised probes at differing points during the first half of the forward pass.}}
    \label{fig:unsupervised_layers}
\end{figure*}

We train supervised and unsupervised probes on models from the Gemma 2 and Llama 3.1 families on the MT-Bench dataset by harvesting activations on contrasting tokens at \textit{all} layers of a given model. The performance of the downstream trained probes are compared in Figure~\ref{fig:supervised_layers} (supervised probes) and Figure~\ref{fig:unsupervised_layers} (unsupervised probes). 

A comparison of the two figures is striking, and reveals some of the key differences between supervised and unsupervised linear probes. For both, average performance is poorer when probes are trained on the first half of a given model than the second. For supervised probes, the increase in performance is smooth: by around 40-50\% of the way through the forward pass, they are able to learn as best they can for the given task. 

In contrast, we see a discontinuity in unsupervised probe performance. This discontinuity appears after different numbers of layers depending on the model, but we note the larger the model the earlier it appears (for a given model family). This discontinuity sees performance jump from an F1 score of roughly 0.5 (a balance between precision and recall, moderately better than random classification) to a maximum of roughly 0.8.

Taking our results in Section~\ref{subsec:generalisation} at face value, we hypothesize the main reason for this difference is the salience of the desired feature. Supervised probes are in a sense reflecting a quality of the latent space itself, and how easy/difficult it may be to identify any given feature within this space. Unsupervised probes, by design, rely on the assumption that the desired feature is the \textit{most salient} of the contrast pair differences, rather than the existence of the feature at all.

The results in Figure~\ref{fig:unsupervised_layers} suggest that in larger models this quality of salience is realised earlier in the forward pass, perhaps due to higher representational capacity. 

Nonetheless, our decision to harvest activations at the last layer of a given model appears justified, as performance in both Figure~\ref{fig:supervised_layers} and Figure~\ref{fig:unsupervised_layers} remains at its best through the last layer. For practitioners, this is particularly convenient as extraction of the last hidden state of a given model is more easily facilitated in common open-LLM frameworks than earlier layers.

\section{Causal Analysis Of Probe Directions} \label{app:causal}

Within the context of concept-based interpretability of LLMs, the term \textit{feature} is ill-defined. Specifically, it is unclear what exactly constitutes a ``true'' feature of a given model. One possible definition is causal in nature: were the ablation of a given feature representation to result in a model unable to represent said feature, it is in some sense ``true'' and causally relevant during the forward pass. This idea has been used to investigate and ``steer'' LLM features in previous works such as \citet{arditi2024refusal} and \citet{rimsky-etal-2024-steering}, and we follow a similar approach here to investigate the extent to which the features identified by our supervised and unsupervised probes are ``true''.

We repeat the prompting experiments performed using the MT-Bench dataset with models from the Gemma 2 and Llama 3.1 families. However, for each model, we orthogonalize the last token's embedding against either the supervised or unsupervised probe directions identified before, at all layers during the forward pass. That is, after each decoder block and the final layer normalisation, we perform the vector rejection,
\begin{align*}
    x' = x - \frac{x \cdot p}{p \cdot p}p,
\end{align*}
replacing the original hidden state vector $x$ with $x'$ given the probe direction $p$, meaning the model's computational operations are never permitted to write information along this probe direction. The difference in evaluator performances (from the baseline un-steered model) are shown in Table~\ref{tab:steering}.

\begin{table*}[t]
    \centering
    \resizebox{\linewidth}{!}{
        \begin{tabular}{|>{\columncolor{customgray}}c|>{\columncolor{customgray}}c|>{\columncolor{customgray}}c|>{\columncolor{customgray}}c|>{\columncolor{customgray}}c|>{\columncolor{customgray}}c|}
        \hline
        $\Delta$ F1 Score & Gemma 2 2B & Gemma 2 9B & Gemma 2 27B & Llama 3.1 8B & Llama 3.1 70B \\
        \hline
        Supervised & -0.00 & 0.01 & 0.00 & 0.00 & 0.03 \\
        Unsupervised & -0.01 & 0.01 & 0.00 & 0.01 & 0.03 \\
        \hline
        \end{tabular}
    }
    \vspace{0.5em}
    \caption{Change ($\Delta$ F1) in evaluator performance on the MT-Bench dataset following the ablation of a given probe direction during the forward pass. For all models tested, we see neglibible change in the model's capability when it is unable to write information against the probe direction, suggesting these directions are not causally relevant for evaluation.}
    \label{tab:steering}
\end{table*}

We see negligible change in evaluator performance regardless of probe type, and consider this evidence against the hypothesis that the features classifying probes identify are true features used by a given language model during LLM-as-a-Judge evaluation. 

This may be due to the nature of high-dimensional space: it is likely there are several high-performing linear classifiers for such a task, and our probes are only ever able to identify one. 

It may also be the case that features used for the \textit{expression} of a belief are different from those used to assess belief in a token \textit{already generated}. Note in the contrast pair setup, activations are harvested at the contrasting token position, as opposed to the token before. It may well be that our probes, in particular our unsupervised probes, are identifying features related to a model's belief in its own utterances, rather than features related to evaluation itself. This raises intriguing questions regarding how realistic off-policy vs on-policy experiments with LLMs are, and we would be excited to see this explored in future work.

\section{Additional Results For Our Ablation Study} \label{app:ablation}

We repeat the experiments performed in Section~\ref{subsec:ablations} on Gemma 2 2B, 9B, 27B, and Llama 3.1 8B. Results are consistent with our tests on Llama 3.1 70B in that probes are, in general, more robust to adversarial prompting strategies than generation-based inference. Note this is particularly apparent with the smallest model tested (Gemma 2 2B).

\begin{figure*}[]
    \centering
    \includegraphics[width=0.7\textwidth]{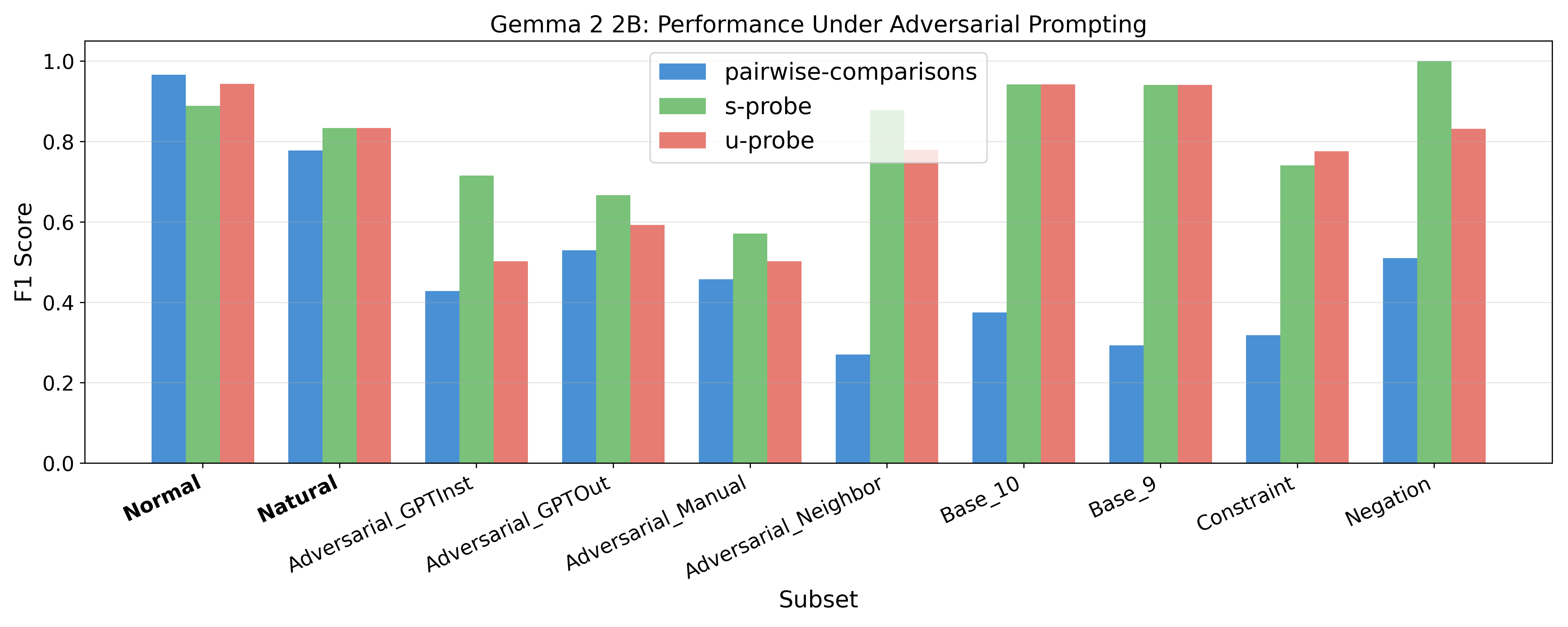}
    \caption{Performance of classifying probes and standard prompting for Gemma 2 2B on the LLMBar dataset.}
    \label{fig:ablation_gemma2b}
\end{figure*}

\begin{figure*}[]
    \centering
    \includegraphics[width=0.7\textwidth]{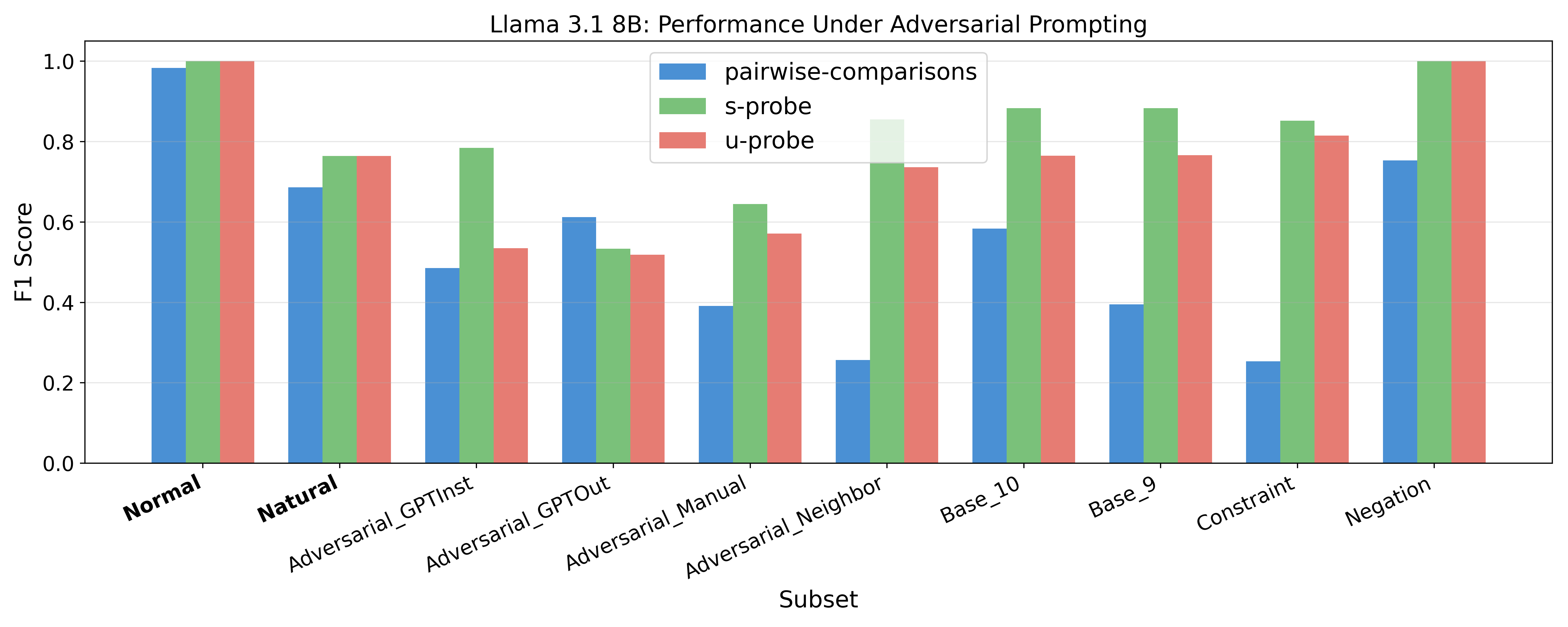}
    \caption{Performance of classifying probes and standard prompting for Llama 3.1 8B on the LLMBar dataset.}
    \label{fig:ablation_llama8b}
\end{figure*}

Results are shown in Figure~\ref{fig:ablation_gemma2b} through to Figure~\ref{fig:ablation_llama8b}.

\begin{figure*}[]
    \centering
    \includegraphics[width=0.7\textwidth]{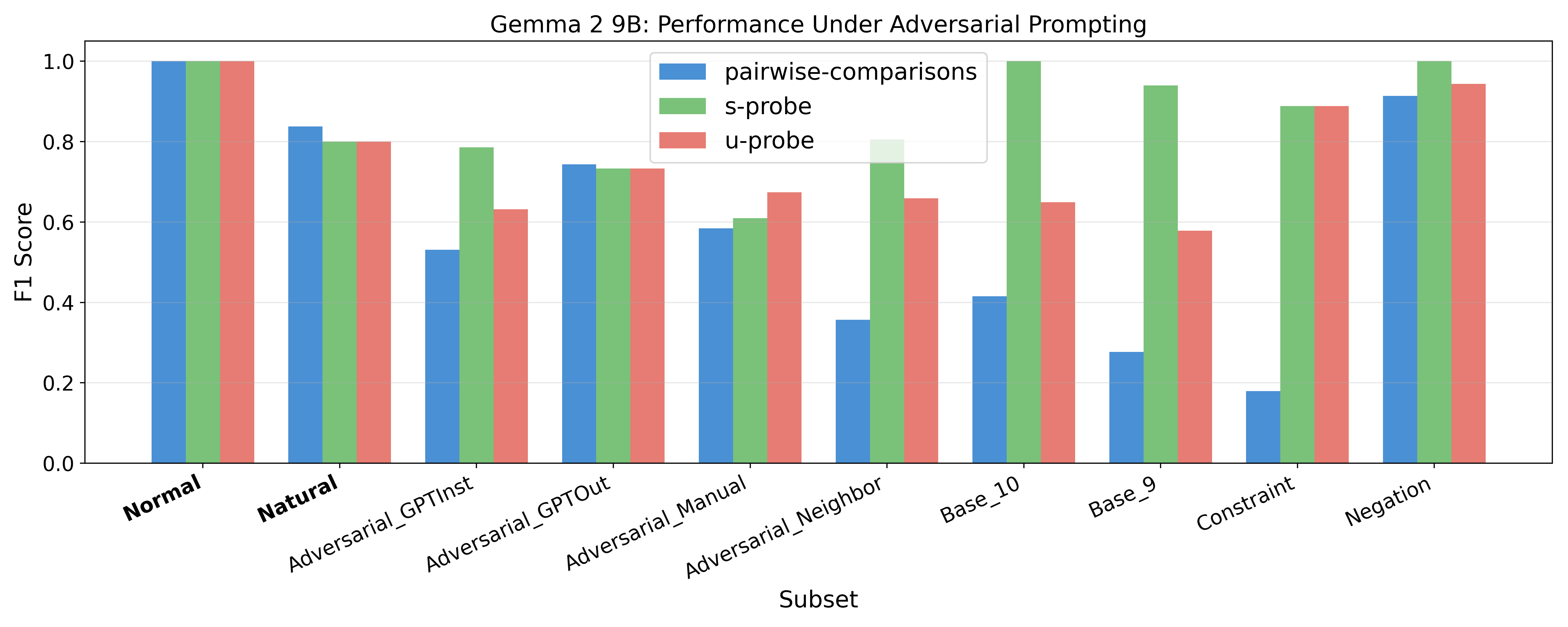}
    \caption{Performance of classifying probes and standard prompting for Gemma 2 9B on the LLMBar dataset.}
    \label{fig:ablation_gemma9b}
\end{figure*}

\begin{figure*}[]
    \centering
    \includegraphics[width=0.7\textwidth]{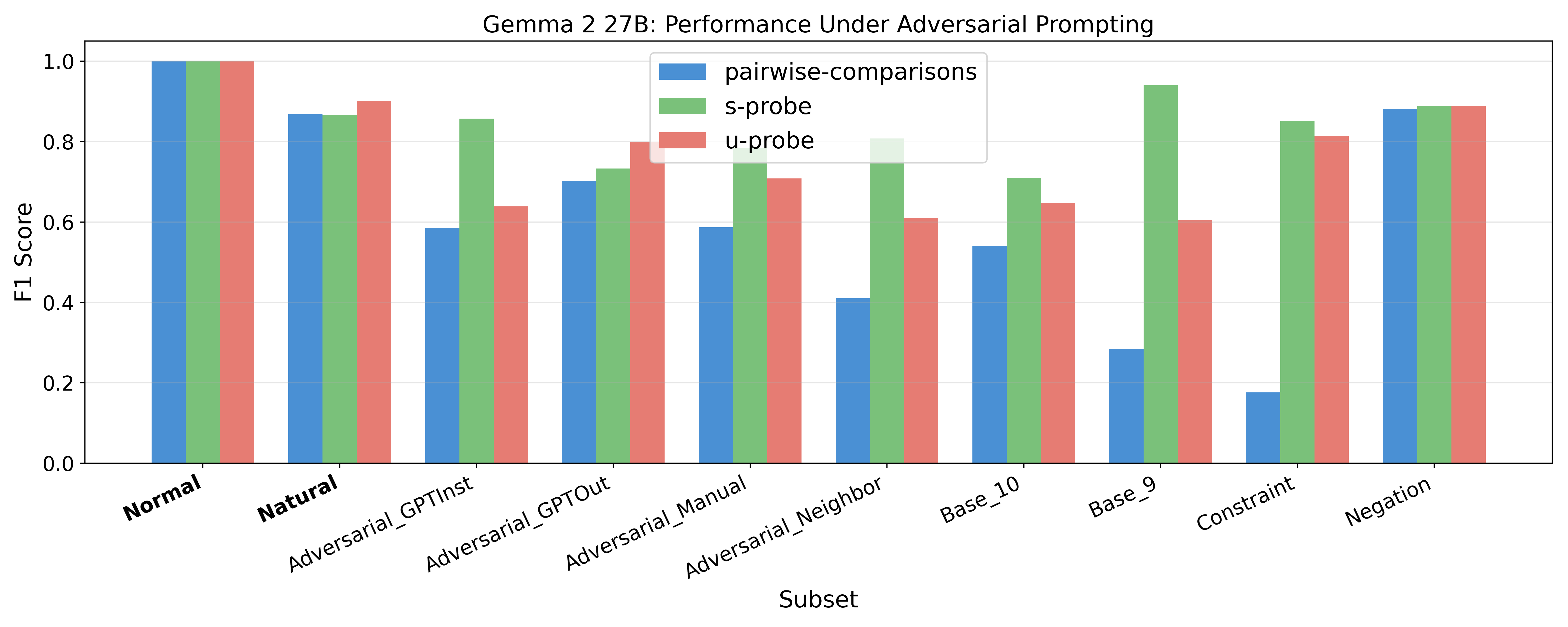}
    \caption{Performance of classifying probes and standard prompting for Gemma 2 27B on the LLMBar dataset.}
    \label{fig:ablation_gemma27b}
\end{figure*}

\section{Details Of Supervised Finetuning Experiments} \label{app:finetuning}

For the experiments in Section~\ref{subsec:supervised_probes} in which we compare supervised probe performance with finetuning for models Gemma 2 2B, 9B, and 27B. We use the OpenRLHF \citep{hu2024openrlhf} library. For LoRA \citep{hu2022lora} finetuning we use a rank of 64 and $\alpha$ of 64, targeting all modules. In all cases we train on a dataset of 5000 randomly chosen samples for one epoch. Full training configs will be available in our code repository to be published.

\section{Full Prompts for All Datasets} \label{app:prompts}

\subsection{Text Quality Datasets}

The text quality datasets NEWSROOM \citep{grusky-etal-2018-newsroom}, SummEval \citep{fabbri2020summeval}, and HANNA \citep{chhun2024do} all present the task of the evaluation of generated text on high-level, abstract features or aspects. Each original dataset includes descriptions of these features, which we use in our evaluator prompts for additional context. These descriptions are listed in Table~\ref{tab:text-quality-descriptions}.

\begin{table*}[t]
    \centering
    \resizebox{\linewidth}{!}{
    \begin{tabular}{|l | c | p{10cm}|}
        \hline
        \textbf{Dataset} & \textbf{Aspect} & \textbf{Description} \\
        \hline
        \multirow{4}{*}[-1.5em]{\rotatebox[origin=c]{90}{NEWSROOM}} 
        & Informativeness & Informativeness is how well a summary of an article captures the key points of the article. \\
        & Relevance & The details provided in a relevant summary of an article are consistent with details in the article. \\
        & Fluency & In a fluent summary of an article the individual sentences are well-written and grammatical. \\
        & Coherence & In a coherent summary of an article the phrases and sentences fit together and make sense collectively. \\
        \hline
        \multirow{4}{*}[-6em]{\rotatebox[origin=c]{90}{SummEval}} 
        & Coherence & Coherence is the collective quality of all sentences. A coherent summary of a source should be well-structured and well-organized. It should not be a heap of related information, but should build from sentence to sentence to a coherent body of information about the source. \\
        & Consistency & Consistency is the factual alignment between a summary and summarized source. A coherent summary contains only statements that are entailed by the source document. \\
        & Fluency & Fluency is the quality of individual sentences. A fluent summary of a source should have no formatting problems, capitalization errors or obviously ungrammatical sentences (e.g., fragments, missing components) that make the text difficult to read. \\
        & Relevance & Relevance is the selection of important content from a source. A relevant summary should include only important information from the source document. \\
        \hline
        \multirow{6}{*}[-0.75em]{\rotatebox[origin=c]{90}{HANNA}} 
        & Relevance & A relevant story matches its prompt. \\
        & Coherence & A coherent story makes sense. \\
        & Empathy & An empathetic story allows the reader to understand the character's emotions. \\
        & Surprise & A surprising story has a surprising end. \\
        & Engagement & An engaging story allows the reader to engage with it. \\
        & Complexity & A complex story is elaborate. \\
        \hline
    \end{tabular}
    }
    \vspace{0.5em}
    \caption{Text descriptions of high-level abstract features/aspects in all text quality datasets. These are provided in prompts for additional context.}
    \label{tab:text-quality-descriptions}
\end{table*}

The prompt formats for the NEWSROOM and SummEval datasets follow a very similar structure, as both assess the same exact task: a news article \texttt{CONTEXT} is provided with given summaries (\texttt{ITEM}s). We include the relevant \texttt{DESCRIPTION} according to the \texttt{ASPECT} under study, consulting Table~\ref{tab:text-quality-descriptions}. For the direct-scoring setting, the prompt template used for NEWSROOM is:

\begin{verbatim}
Consider the following article and summary:
Article: {CONTEXT}
Summary: {ITEM}
{DESCRIPTION} Rate the {ASPECT} of this 
summary from 1 to 5, where 1 represents 
very low {ASPECT}, and 5 represents 
excellent {ASPECT}. Responses must be a 
single score.
\end{verbatim}

For SummEval, the template is changed slightly, to match the original dataset and paper:

\begin{verbatim}
Consider the following source and summary:
Source: {CONTEXT}
Summary: {ITEM}
{DESCRIPTION} Rate the {ASPECT} of this 
summary from 1 to 5, where 1 represents 
very low {ASPECT}, and 5 represents 
excellent {ASPECT}. Responses must be a 
single score.
\end{verbatim}

For pairwise comparisons, we follow a very similar template. For NEWSROOM:

\begin{verbatim}
Consider the following article:
Article: {CONTEXT}
Below are two summaries of the above 
article:
Summary 1: {ITEM1}
Summary 2: {ITEM2}
{DESCRIPTION} Which summary is more 
{ASPECT}? Responses must be a single 
choice.
\end{verbatim}

And for SummEval:

\begin{verbatim}
Consider the following source:
Source: {CONTEXT}
Below are two summaries of the above 
source:
Summary 1: {ITEM1}
Summary 2: {ITEM2}
{DESCRIPTION} Which summary is more 
{ASPECT}? Responses must be a single 
choice.
\end{verbatim}

For the HANNA dataset, we evaluate stories generated from story-prompts. The above template is therefore adjusted slightly. For direct-scoring:

\begin{verbatim}
Consider the following prompt and story:
Prompt: {CONTEXT}
Story: {ITEM}
{DESCRIPTION} Rate the {ASPECT} of this 
story from 1 to 5, where 1 represents 
very low {ASPECT}, and 5 represents 
excellent {ASPECT}. Responses must be a 
single score.
\end{verbatim}

And for pairwise comparisons:

\begin{verbatim}
Consider the following prompt:
Prompt: {CONTEXT}
Below are two stories inspired 
by the above prompt:
Story 1: {ITEM1}
Story 2: {ITEM2}
{DESCRIPTION} Which story is more 
{ASPECT}? Responses must be a single 
choice.
\end{verbatim}

\subsection{Common Sense Reasoning Datasets}

For the ROCStories \citep{mostafazadeh-etal-2016-corpus}, MCTACO citep{zhou-etal-2019-going}, and CateRS \citep{mostafazadeh-etal-2016-caters} datasets, the task is formatted only as one of pairwise comparisons. Additionally, in all cases the evaluator must pick the more sensible option, so all prompt templates are very similar. For ROCStories:

\begin{verbatim}
Consider the following short story:
Story: {STORY}
Below are two statements:
Statement 1: {STATEMENT1}
Statement 2: {STATEMENT2}
Considering the context of the above 
story, which statement is more 
consistent? Responses must be a single 
choice.
\end{verbatim}

With each \texttt{STORY} and \texttt{STATEMENT}s obtained from the dataset directly. 

Similarly for MCTACO:

\begin{verbatim}
Consider the following passage:
Passage: {PASSAGE}
Below is a question regarding 
the above passage:
Question: {QUESTION}
Choice 1: {CHOICE1}
Choice 2: {CHOICE2}
Which answer is more sensible? 
Responses must be a single choice.
\end{verbatim}

This dataset assesses common sense reasoning through a specific \texttt{QUESTION} for each \texttt{PASSAGE}.

Lastly, for CaTeRS:

\begin{verbatim}
The following list of statements 
form a story, however they are 
unordered:
Unordered Statements: {UNORDERED}
Below are two statements from this 
list:
Statement 1: {STATEMENT1}
Statement 2: {STATEMENT2}
Determine the correct order of the 
above statements - which statement 
appears before the other? Responses 
must be a single choice.
\end{verbatim}

This dataset includes lists of unordered statements, with the pairwise comparison task set up of identifying the correct ordering of two such statements, thereby assessing temporal understanding.

\section{Probe Performance by Dataset} \label{app:full_probes}

\begin{figure*}[!h]
    \centering
    \includegraphics[width=0.7\textwidth]{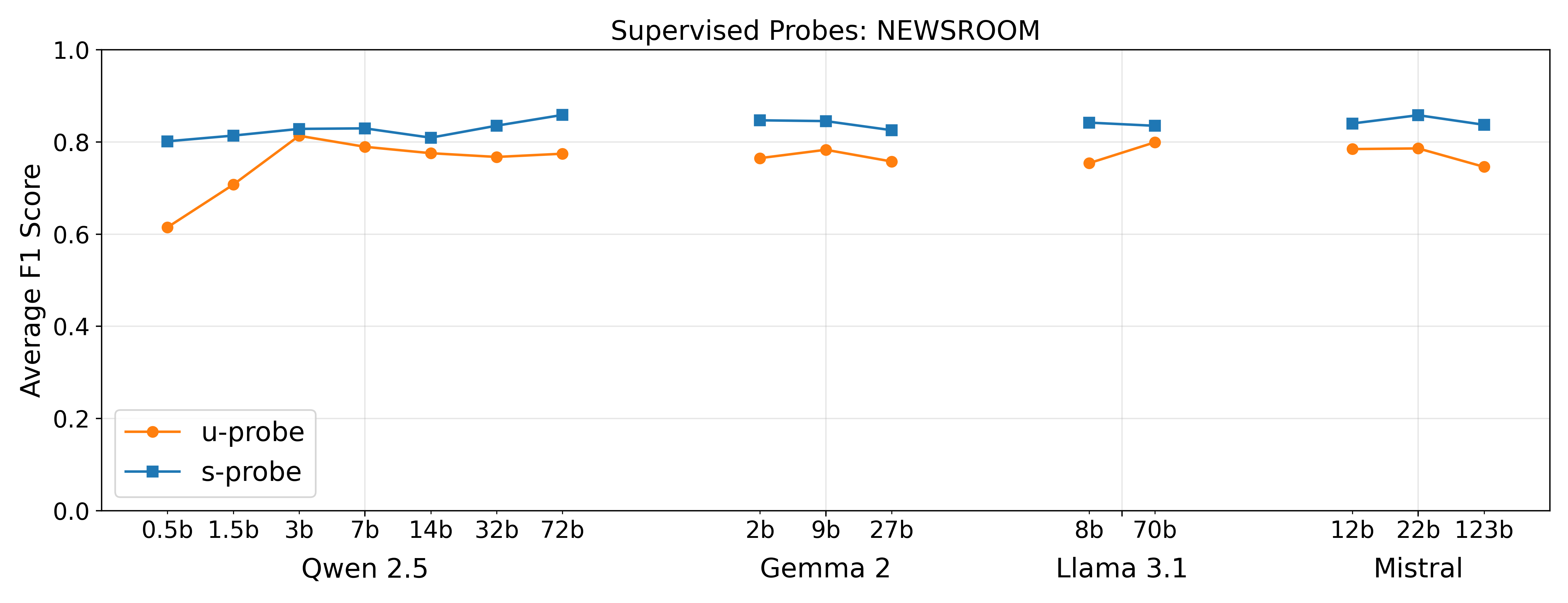}
    \caption{Supervised probe performance on the NEWSROOM dataset.}
    \label{fig:supervised_newsroom}
\end{figure*}

\begin{figure*}[!h]
    \centering
    \includegraphics[width=0.7\textwidth]{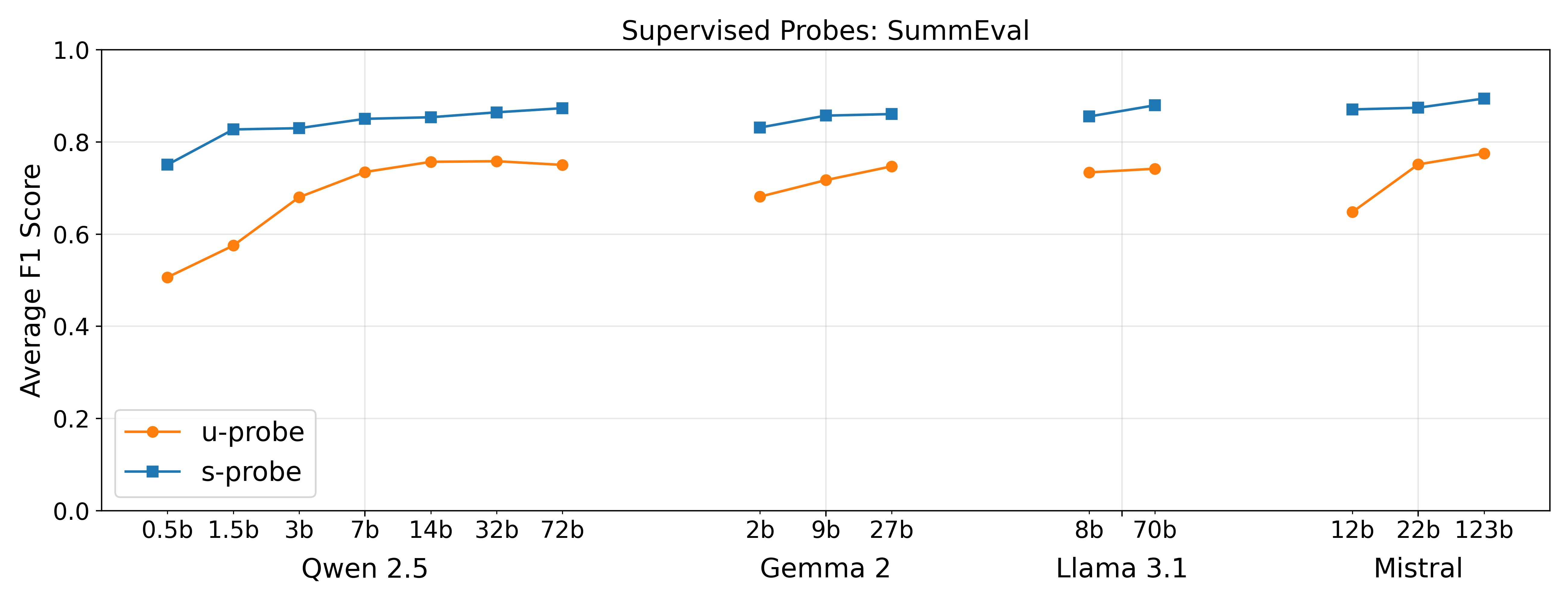}
    \caption{Supervised probe performance on the SummEval dataset.}
    \label{fig:supervised_summeval}
\end{figure*}

\begin{figure*}[!h]
    \centering
    \includegraphics[width=0.7\textwidth]{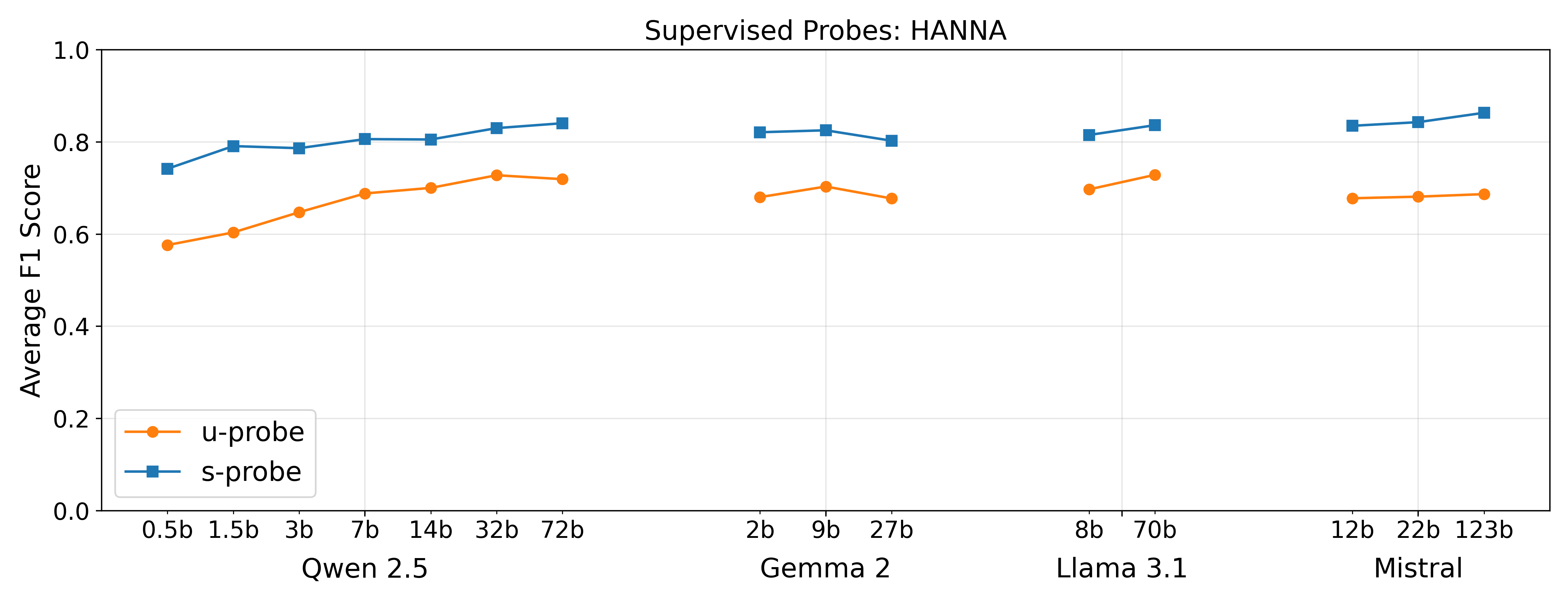}
    \caption{Supervised probe performance on the HANNA dataset.}
    \label{fig:supervised_hanna}
\end{figure*}

\begin{figure*}[!h]
    \centering
    \includegraphics[width=0.7\textwidth]{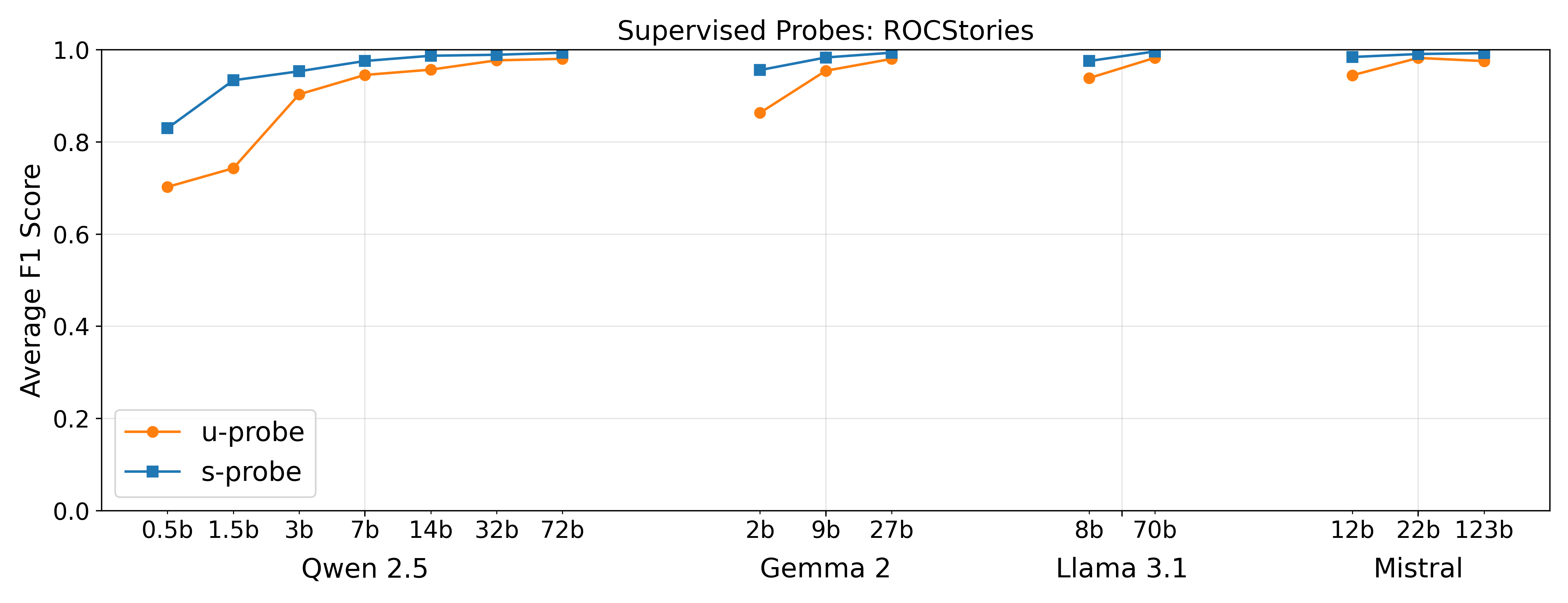}
    \caption{Supervised probe performance on the ROCStories dataset.}
    \label{fig:supervised_rocstories}
\end{figure*}

\begin{figure*}[!h]
    \centering
    \includegraphics[width=0.7\textwidth]{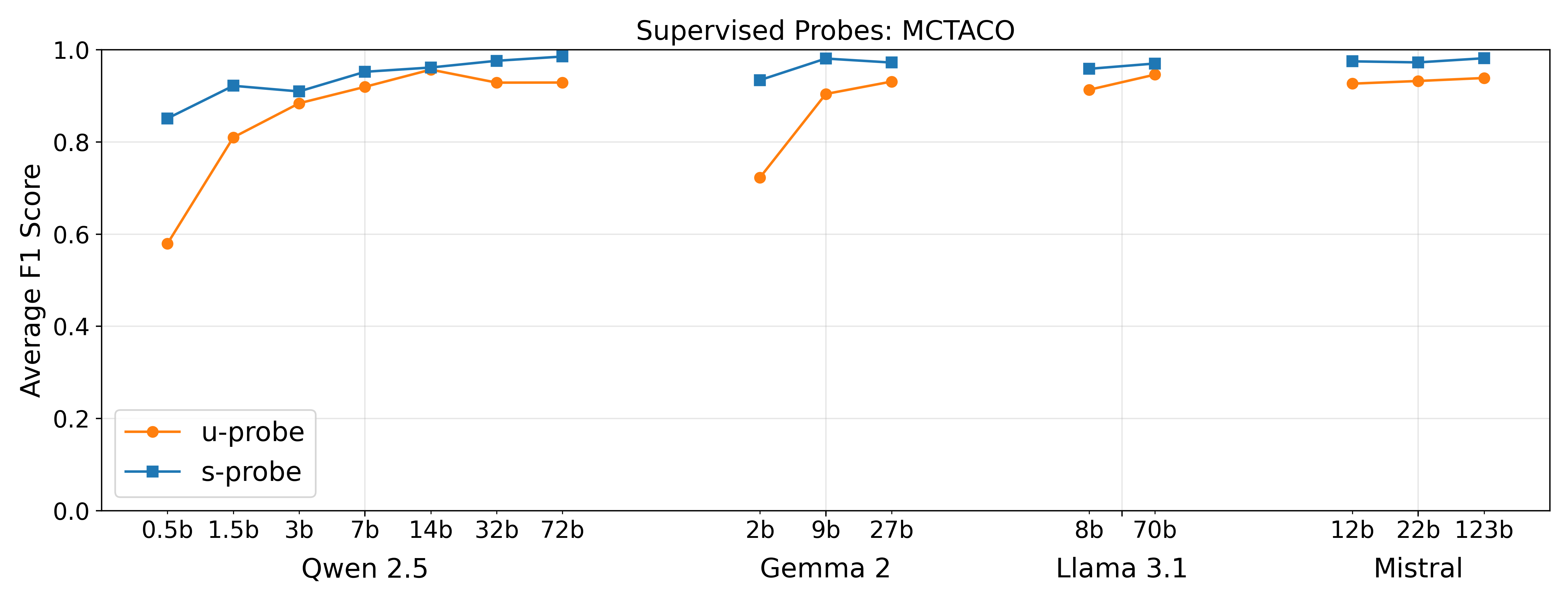}
    \caption{Supervised probe performance on the MCTACO dataset.}
    \label{fig:supervised_mctaco}
\end{figure*}

\begin{figure*}[!h]
    \centering
    \includegraphics[width=0.7\textwidth]{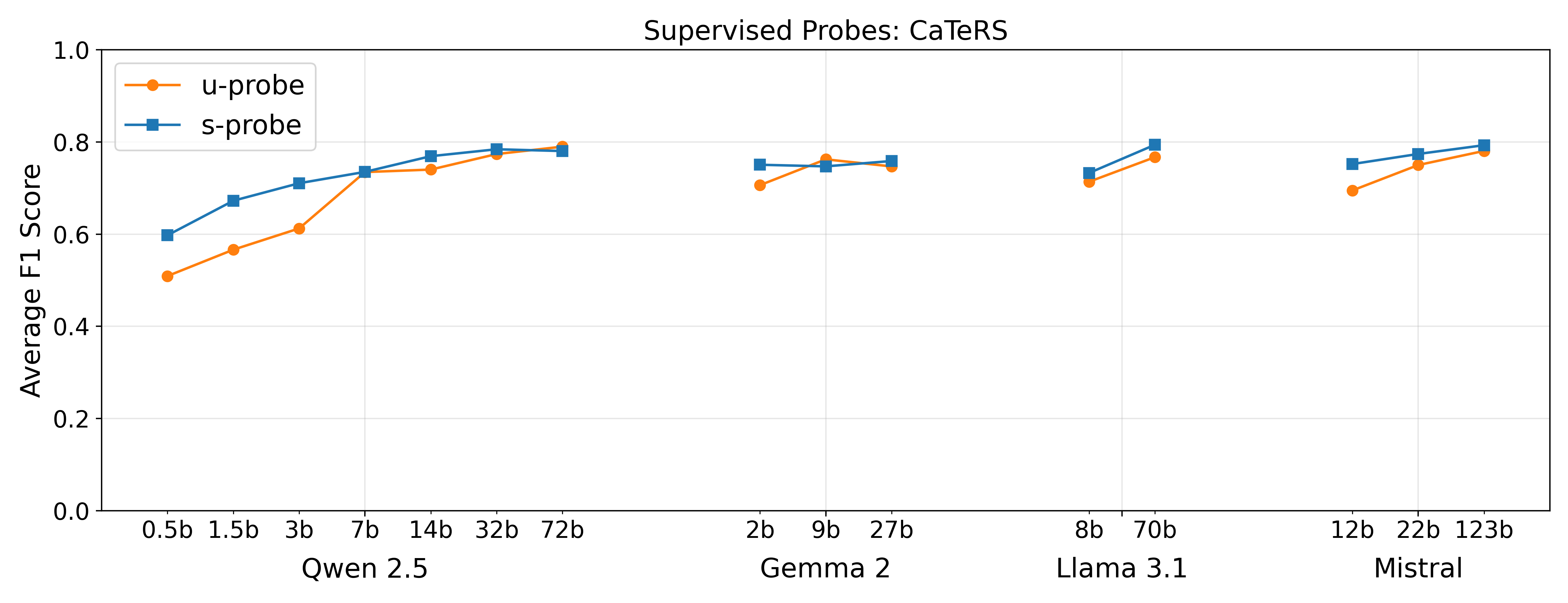}
    \caption{Supervised probe performance on the CaTeRS dataset.}
    \label{fig:supervised_caters}
\end{figure*}

\begin{figure*}[!h]
    \centering
    \includegraphics[width=0.7\textwidth]{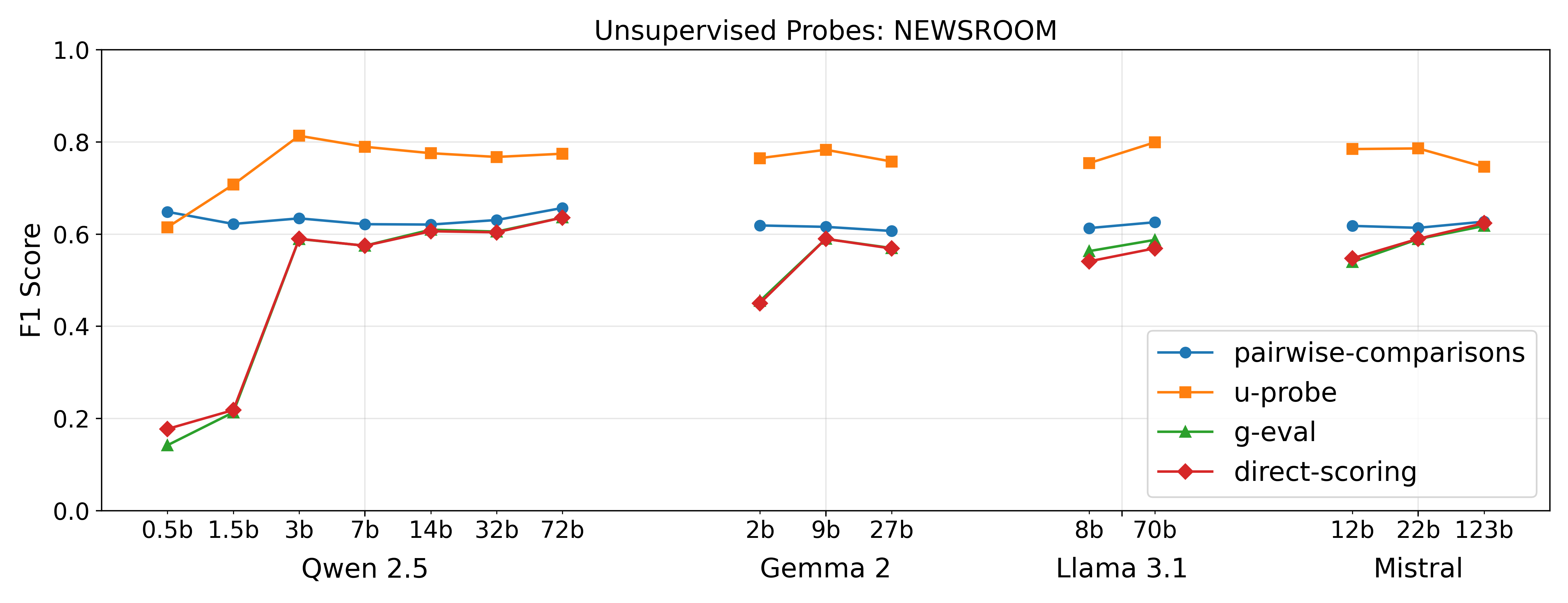}
    \caption{Unsupervised probe performance on the NEWSROOM dataset.}
    \label{fig:unsupervised_newsroom}
\end{figure*}

\begin{figure*}[!h]
    \centering
    \includegraphics[width=0.7\textwidth]{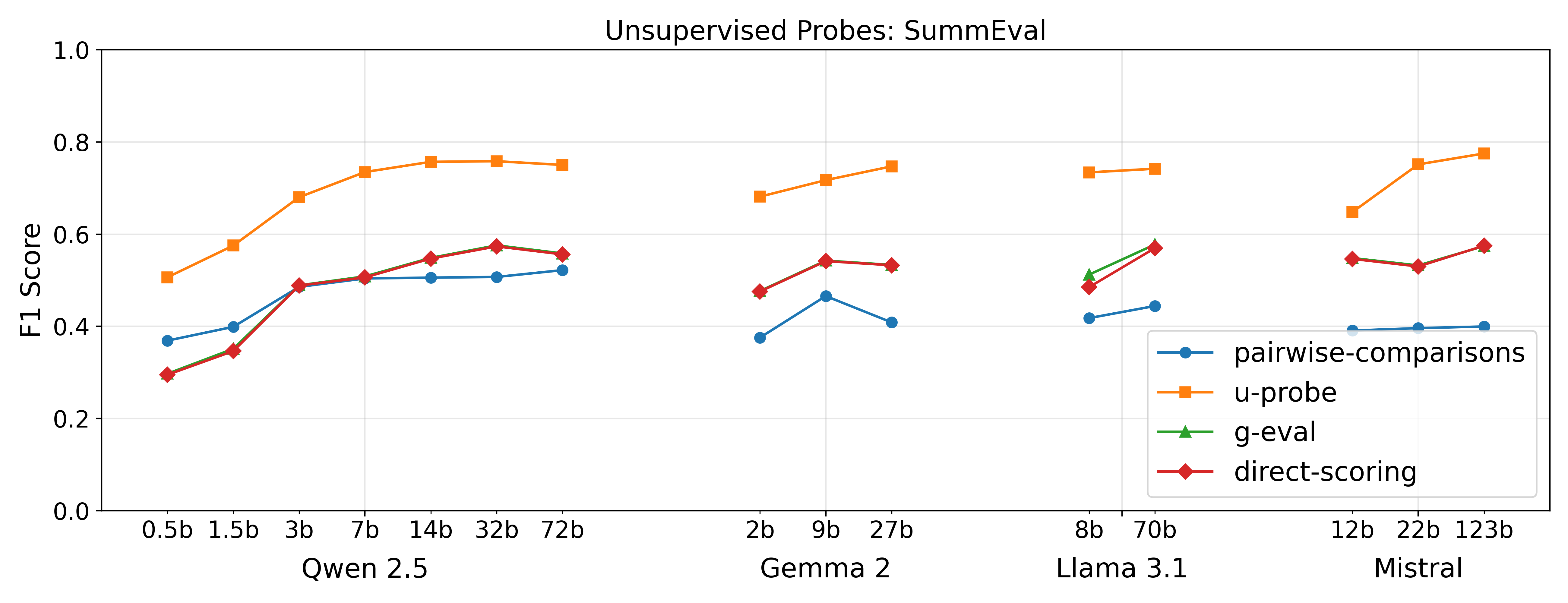}
    \caption{Unsupervised probe performance on the SummEval dataset.}
    \label{fig:unsupervised_summeval}
\end{figure*}

\begin{figure*}[!h]
    \centering
    \includegraphics[width=0.7\textwidth]{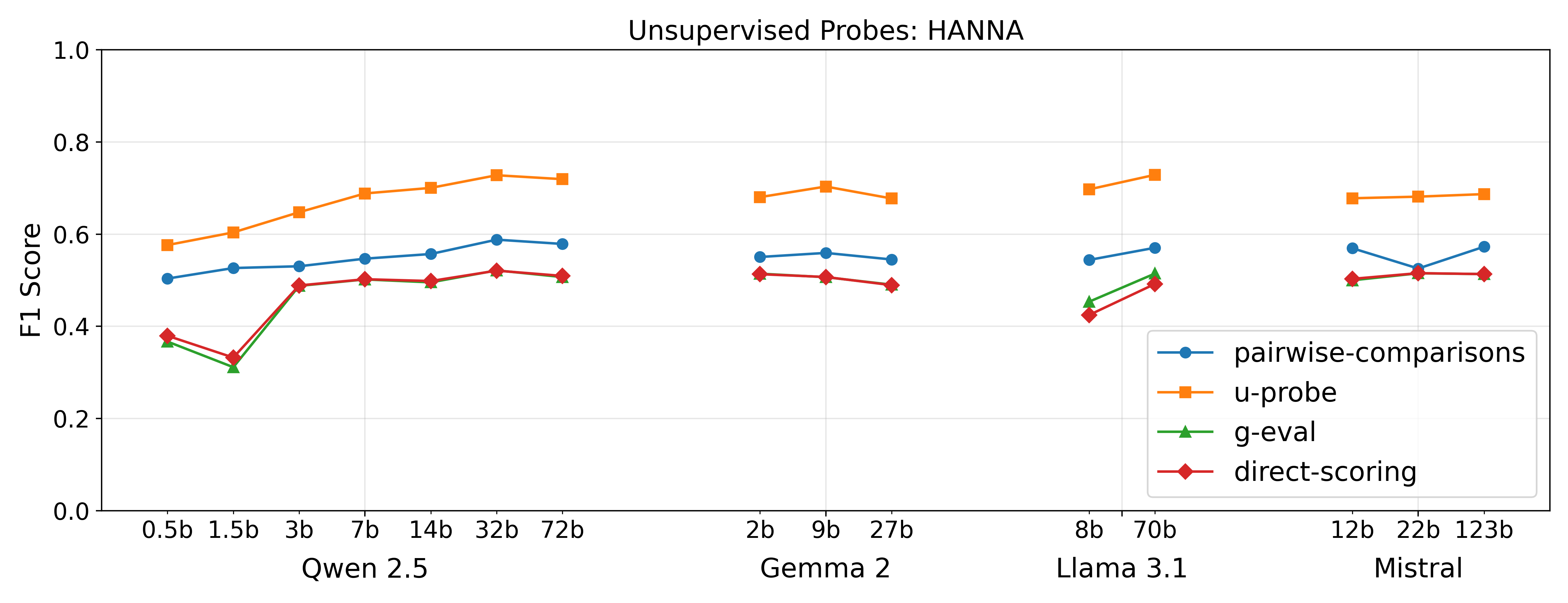}
    \caption{Unsupervised probe performance on the HANNA dataset.}
    \label{fig:unsupervised_hanna}
\end{figure*}

\begin{figure*}[!h]
    \centering
    \includegraphics[width=0.7\textwidth]{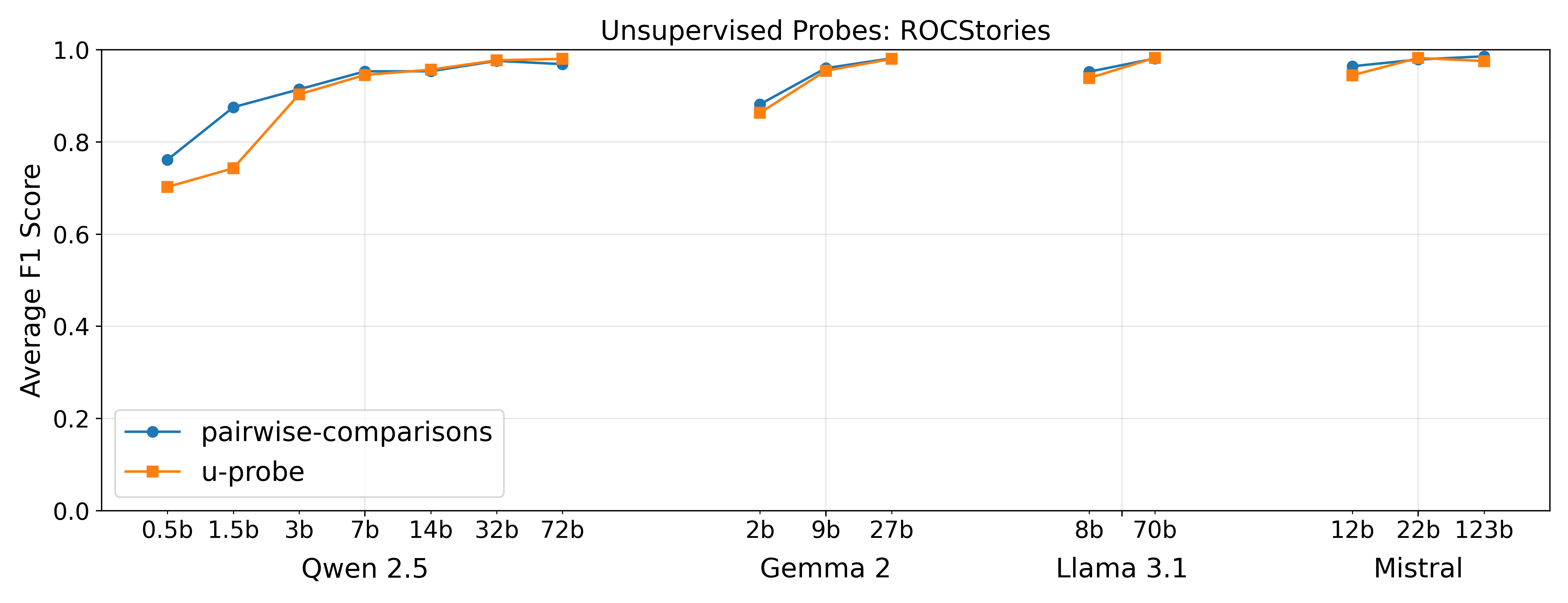}
    \caption{Unsupervised probe performance on the ROCStories dataset.}
    \label{fig:unsupervised_rocstories}
\end{figure*}

\begin{figure*}[!h]
    \centering
    \includegraphics[width=0.7\textwidth]{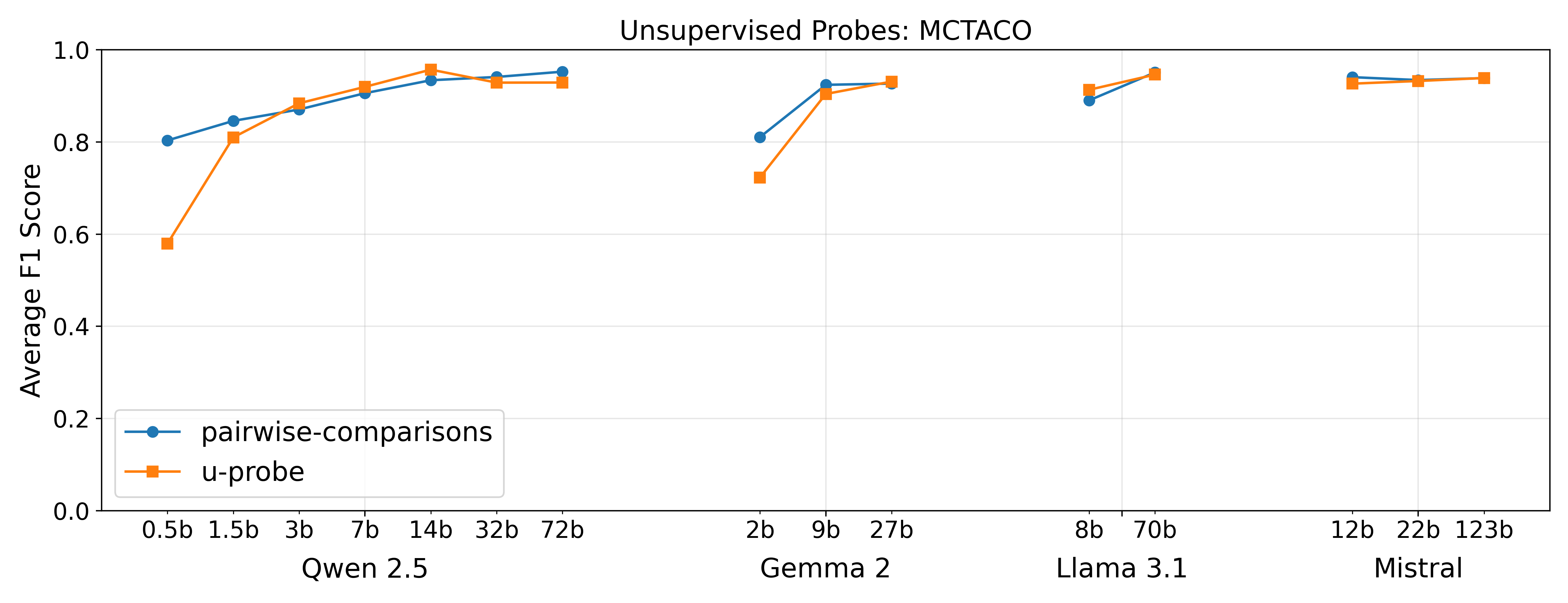}
    \caption{Unsupervised probe performance on the MCTACO dataset.}
    \label{fig:unsupervised_mctaco}
\end{figure*}

\begin{figure*}[!h]
    \centering
    \includegraphics[width=0.7\textwidth]{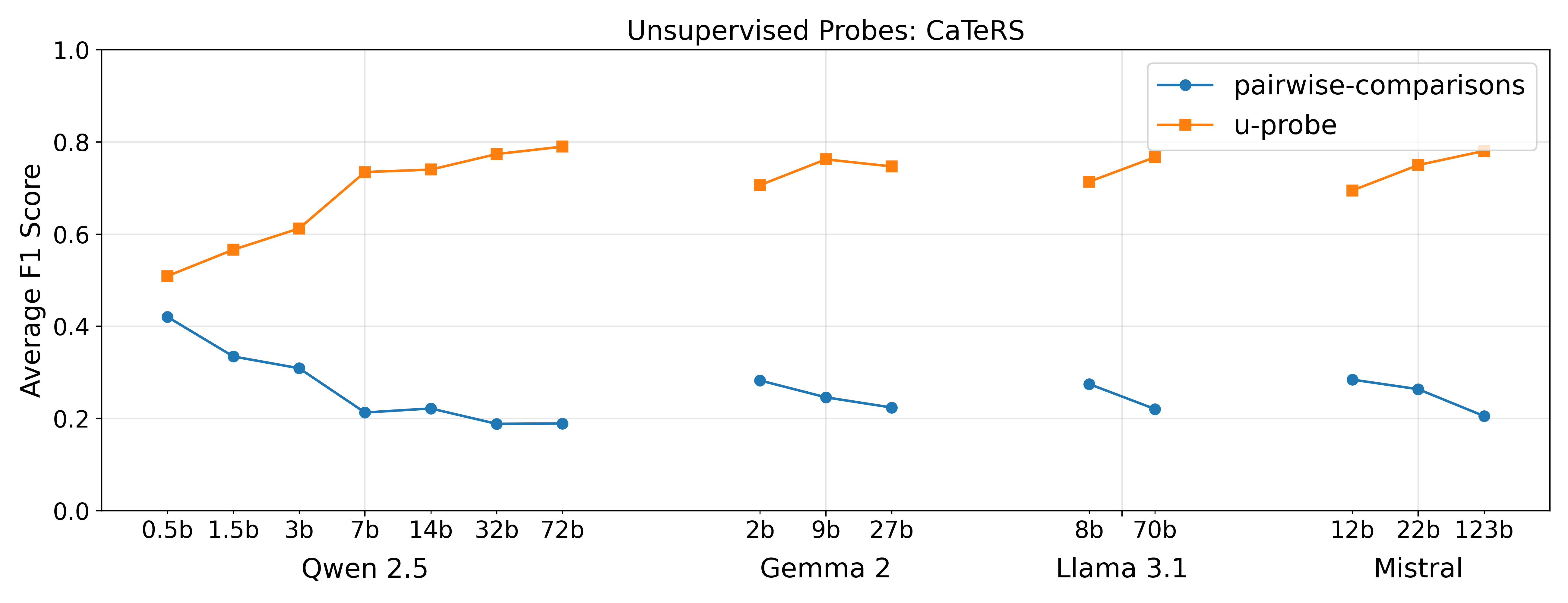}
    \caption{Unsupervised probe performance on the CaTers dataset.}
    \label{fig:unsupervised_caters}
\end{figure*}

We present the performance of supervised (Figure~\ref{fig:supervised_newsroom} to Figure~\ref{fig:supervised_caters}) and unsupervised (Figure~\ref{fig:unsupervised_newsroom} to Figure~\ref{fig:unsupervised_caters}) probes on all constituent datasets of the text quality (NEWSROOM \citep{grusky-etal-2018-newsroom}, SummEval \citep{fabbri2020summeval}, HANNA \citep{chhun2024do}) and common sense reasoning (ROCStories \citep{mostafazadeh-etal-2016-corpus}, MCTACO \citep{zhou-etal-2019-going}, CaTeRS \citep{mostafazadeh-etal-2016-caters}) tasks examined in Section~\ref{sec:results}.

\end{document}